\pdfoutput=1

\documentclass[11pt]{article}

\usepackage[]{acl}

\usepackage{times}
\usepackage{latexsym}
\usepackage{colortbl}
\usepackage{xcolor}
\usepackage{array}
\usepackage{makecell}
\usepackage{tabularx}
\usepackage{url}
\usepackage{longtable}
\usepackage{amssymb} 
\usepackage{amsmath}  
\usepackage{multirow} 
\usepackage{booktabs} 
\usepackage{authblk}
\usepackage{lipsum}   
\usepackage{pifont}
\usepackage[T1]{fontenc}

\newcommand{\tabincell}[2]{\begin{tabular}{@{}#1@{}}#2\end{tabular}}
\usepackage[utf8]{inputenc}

\usepackage{microtype}

\usepackage{inconsolata}

\usepackage{graphicx}

\title{Towards Lightweight, Adaptive and Attribute-Aware Multi-Aspect Controllable Text Generation with Large Language Models}

\author{
    {\bf Chenyu Zhu$^\dagger$} \quad
    {\bf Yefeng Liu$^\dagger$}\thanks{Corresponding author.} \quad
    {\bf Chenyang Lyu$^\dagger$} \quad
    {\bf Xue Yang$^\dagger$} \quad
    {\bf Guanhua Chen$^\ddagger$} \quad \\
    {\bf Longyue Wang$^\dagger$} \quad
    {\bf Weihua Luo$^\dagger$} \quad
    {\bf Kaifu Zhang$^\dagger$} \\
    $^\dagger$Alibaba International Digital Commerce \\
    $^\ddagger$Southern University of Science and Technology \\
    {\normalsize cyzhu@zju.edu.cn,~~fengzhi.lyf@alibaba-inc.com}
}

\begin{document}
\maketitle
\begin{abstract}
Multi-aspect controllable text generation aims to control text generation in attributes from multiple aspects, making it a complex but powerful task in natural language processing. Supervised fine-tuning methods are often employed for this task due to their simplicity and effectiveness. However, they still have some limitations: low rank adaptation (LoRA) only fine-tunes a few parameters and has suboptimal control effects, while full fine-tuning (FFT) requires significant computational resources and is susceptible to overfitting, particularly when data is limited. Moreover, existing works typically train multi-aspect controllable text generation models using only single-aspect annotated data, which results in discrepancies in data distribution; at the same time, accurately generating text with specific attributes is a challenge that requires strong attribute-aware capabilities. To address these limitations, we propose a lightweight, adaptive and attribute-aware framework for multi-aspect controllable text generation. Our framework can dynamically adjust model parameters according to different aspects of data to achieve controllable text generation, aiming to optimize performance across multiple aspects. Experimental results show that our framework outperforms other strong baselines, achieves state-of-the-art performance, adapts well to data discrepancies, and is more accurate in attribute perception.

\end{abstract}

\section{Introduction}
Controllable text generation has become an important research direction in natural language processing, enabling the generation of text that adheres to specific attributes or constraints \cite{carlsson-etal-2022-fine,yang-etal-2023-tailor}. This capability is pivotal for personalized content creation, dialogue systems, and data augmentation applications. 
Existing approaches for controllable text generation can be generally divided into three main categories: fine-tuning \cite{keskarCTRL2019,ziegler2020finetuninglanguagemodelshuman}, latent space manipulation \cite{gu-etal-2022-distributional,gu-etal-2023-controllable}, and decoding-time intervention \cite{dathathri2019plug,Li-2022-DiffusionLM}. These approaches aim to steer the generative model toward producing text that aligns with the desired control attributes while maintaining fluency and coherence.

\begin{table}[t]
    \centering
    \resizebox{1\linewidth}{!}{
    \begin{tabular}{cccc}
        \toprule
        \tabincell{c}{\bf Method} & \tabincell{c}{\bf Lightweight} & \tabincell{c}{\bf Adaptive} & \tabincell{c}{\bf Attribute-Aware} \\
        \midrule
        LoRA & \textcolor{blue}{\ding{51}} & \textcolor{red}{\ding{55}} & \textcolor{red}{\ding{55}} \\
        FFT & \textcolor{red}{\ding{55}} & \textcolor{red}{\ding{55}} & \textcolor{red}{\ding{55}} \\
        Ours & \textcolor{blue}{\ding{51}} & \textcolor{blue}{\ding{51}} & \textcolor{blue}{\ding{51}} \\
        \bottomrule
    \end{tabular}
    }
    \caption{Comparison of our method with LoRA and FFT. Our framework is characterized by lightweight implementation, adaptive to data discrepancies, and aware of attributes' differences. The detailed comparison results can be found in Section \ref{sec:main results}.}
    \label{tab:comparison}
\end{table}

Supervised fine-tuning is commonly used in controllable text generation tasks \cite{zeldes2020technicalreportauxiliarytuning,li-liang-2021-prefix}. The model's ability to generate text with specific attributes can be enhanced by fine-tuning the model on datasets with specific attributes. Full fine-tuning, which updates all model parameters, allows the model to better adapt to the specific requirements of the target task. LoRA \cite{hu2022lora} is widely used for its ability to reduce computational resource requirements while maintaining efficient fine-tuning performance. However, both methods still have some limitations: Full fine-tuning requires many computing resources to update all model parameters. Although LoRA is more efficient, its performance tends to be less competitive compared to FFT on most tasks \cite{shuttleworth2024lora}. Additionally, neither method can dynamically adjust control strength during the generation process, making it challenging to handle complex attribute interactions.

Furthermore, acquiring training data containing multiple attribute combinations is challenging. Existing approaches typically rely on datasets annotated for a single aspect, with each training sample representing only one attribute. This imbalance in data distribution often diminishes the richness of multi-aspect semantic expression in the generated text. Moreover, since each aspect generally includes a variety of attributes, accurately generating text with a specific attribute becomes particularly challenging, requiring the model to have robust attribute perception capabilities. Due to inherent structural limitations, LoRA and FFT fail to address these issues effectively.

To alleviate these limitations, we propose a lightweight, adaptive and attribute-aware framework for multi-aspect controllable text generation. Specifically, our framework extends traditional LoRA by integrating multiple LoRA modules, each contributing to the control process. Our approach incorporates a gating function and routing strategy to effectively manage the influence of each LoRA module during controllable text generation. This design allows for dynamic adjustment of control strengths across multiple attributes without extensive retraining or model duplication. In addition, we impose constraints on the hidden states containing attributes to reduce the discrepancies of data and perceive the distribution of attributes, thereby achieving more accurate controllable text generation. Experimental results show that our method outperforms strong baselines in controllable text generation, achieving state-of-the-art performance while enhancing the model’s adaptability and robustness. We compare our framework with LoRA and FFT in Table \ref{tab:comparison}. The main contributions of our framework are as below:
\begin{itemize}
    \item We propose a lightweight, adaptive, and attribute-aware multi-aspect controllable text generation framework, which dynamically adjusts model parameters based on different input aspects.
    \item Our framework reduces data discrepancies and perceives the distribution of attributes, thereby enabling precise, controllable text generation.
    \item We conduct experiments based on several open-source LLMs on a publicly available benchmark, and the experimental results demonstrate that our method achieves state-of-the-art (SOTA) performance with better adaptability and robustness.
\end{itemize}

\section{Related Work}
\paragraph{Fine-Tuning.} Fine-tuning refers to the further updating of a pre-trained model's parameters to adapt it to specific tasks \cite{feng-etal-2023-dunst,zheng-etal-2023-click,kumar-etal-2023-controlled}. DisCup \cite{zhang-song-2022-discup} combines a frozen causal language model with an attribute discriminator to optimize control prompts via unlikelihood training. InstructCTG \cite{pmlr-v202-zhou23g} achieves controllable text generation by converting constraints into a natural language instruction dataset and fine-tuning the language model on an augmented dataset. The fine-tuning approach balances adaptability and resource efficiency, making it a common choice for enhancing model performance on specific tasks.

\paragraph{Latent Space Manipulation.} Latent space manipulation aims to control text generation by adjusting internal model representations \cite{chan2021deep,lu-etal-2023-miracle}. PriorControl \cite{gu-etal-2023-controllable} uses probability density estimation methods in latent space to effectively manage complex attribute distributions through reversible transformations. MAGIC \cite{liu-etal-2024-multi} employs disentangled counterfactual augmentation to handle multi-aspect controllable text generation, aiming to balance attribute correlations during training and enhance attribute control during inference by generating counterfactual features in the attribute latent space. These approaches demonstrate flexibility in controlling generation without altering model architecture.

\begin{figure*}[h]
    \centering
    \includegraphics[scale=0.358]{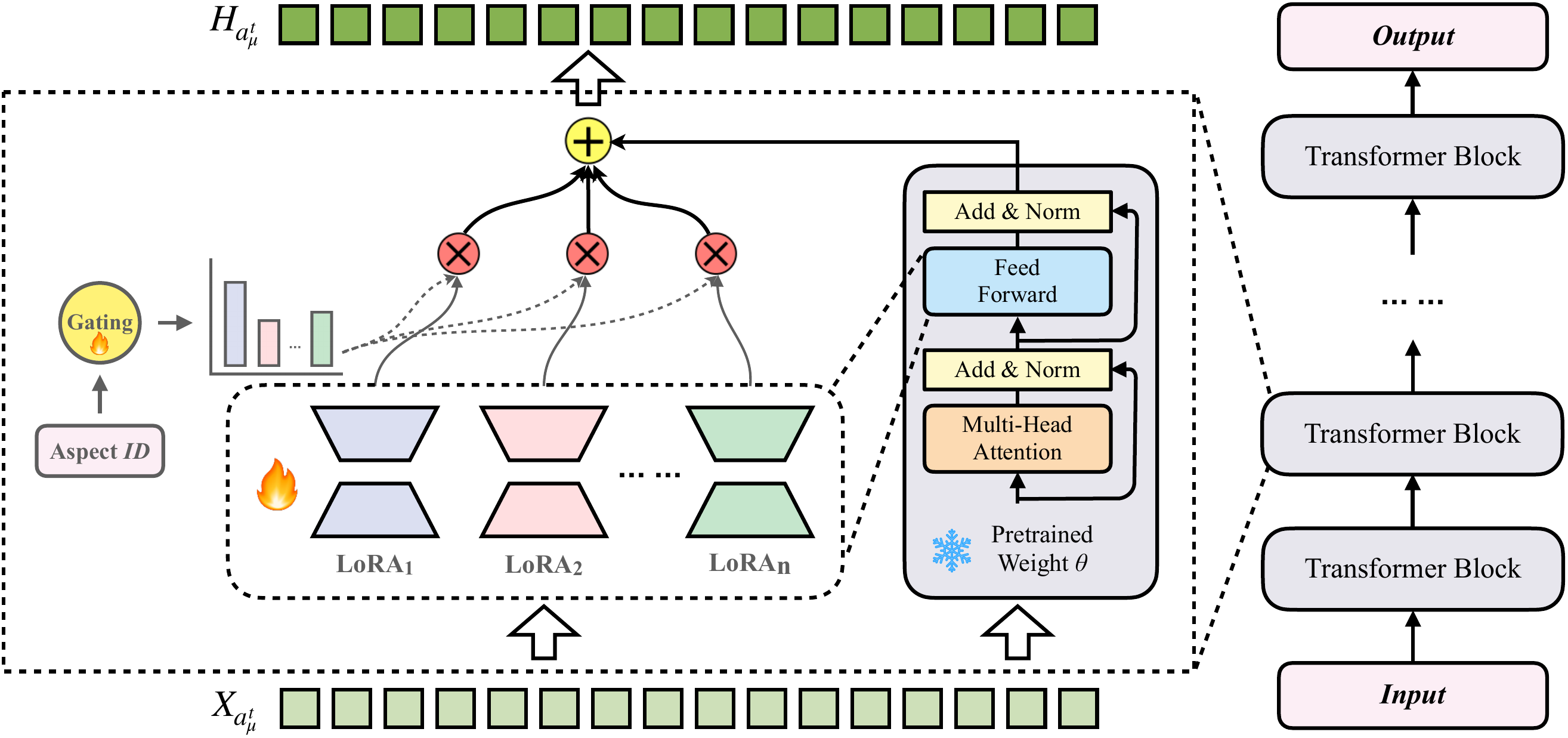}
    \caption{Illustration of our proposed framework. Our framework extends the traditional LoRA by integrating multiple LoRA modules and employs a learnable gating function to dynamically combine multiple LoRA modules. We use the aspect identifier as the input of the gating function to learn unique parameters for each aspect. $X_{a^t_{\mu}}$ represents the input sequence containing attribute $a^t_{\mu}$ and $H_{a^t_{\mu}}$ is the output hidden state. Only the parameters of LoRAs and the gating function are updated during training.}
    \label{fig:1}
\end{figure*}

\paragraph{Decoding-Time Intervention.} Decoding-time intervention controls the attributes of generated text by manipulating the model's logits or probability distribution during generation. PPLM \cite{dathathri2019plug} controls text attributes by iteratively adjusting the hidden layer activations of GPT-2 using the gradient of attribute classifiers. GeDi \cite{krause-etal-2021-gedi-generative} uses a discriminator to guide the language model decoding to calculate the classification probability for each next token, effectively controlling text generation. Air-Decoding \cite{zhong-etal-2023-air} reconstructs attribute distributions to balance the weights between attribute and non-attribute words, effectively generating more fluent and controllable text. Decoding-time intervention enables controllable text generation without training models and is more interpretable.

\section{Formulation}
\paragraph{Task definition.} Let $\mathcal{A} = \{A_1, \ldots, A_N\}$ stands for $N$ aspects. $a^t = \{a^t_1,\ldots, a^t_{|A_t|}\}$ is the set of all attributes in aspect $A_t$, where $|A_t|$ is the number of attributes. We represent text with attribute $a^t_{\mu} \in A_t$ as $T_{a^t_{\mu}}$. The goal of multi-aspect controllable text generation is to generate sentences with multiple attributes from different aspects, such as attribute "joy" from aspect \textit{sentiment} and attribute "sport" from aspect \textit{topic}.

\paragraph{Training samples.} Training samples are instruction following samples; each training sample is consist of four parts: $a^t_{\mu}$ is the control attribute from aspect $A_t$, $I_{a^t_{\mu}}$ is the instruction with attribute $a^t_{\mu}$, $T_{a^t_{\mu}}$ is the target text with attribute $a^t_{\mu}$ and $D_{A_t}$ stands for the identifier of aspect $A_t$. The instructions of all training samples are $I =  {\textstyle \bigcup_{t=1}^{N}}I^t$, where $I^t$ is the instruction set of aspect $A_t$. Identifiers of all aspects are represented as $D = {\textstyle \bigcup_{t=1}^{N}} D_{A_t}$.

\section{Methodology}
In this section, we first introduce the motivation behind our approach and provide an overview of our proposed framework. Subsequently, we present a detailed description of the proposed framework and our training objective.

\subsection{Overview}
In multi-aspect controllable text generation, input instructions typically contain multiple control attributes from different aspects. Therefore, it is necessary to consider the distribution of input instructions dynamically. As illustrated in Figure \ref{fig:1}, our framework takes the Feed-Forward Neural Network (FFN) as an example, where we introduce multiple trainable LoRA modules into the FFN layer to capture diverse knowledge across different controllable text generation tasks. Furthermore, we incorporate a gating function that takes an aspect identifier as input, aiming to learn unique model parameters for each aspect and dynamically combine multiple LoRA modules. At the top level of the transformer block, we impose constraints on hidden states to reduce the discrepancies of data and balance conflicts between attributes.

\subsection{Architecture}
Referring to Figure \ref{fig:1}, consider a Transformer block parameterized by parameters $\theta$, which includes the multi-head attention layer and FFN parameters, and remains constant during training. The trainable set of LoRA modules is represented as $\Delta W = \{\Delta W_1, \ldots, \Delta W_n\}$, where $n$ represents the number of trained LoRAs.

Let $X_{a^t_{\mu}} \in \mathbb{R}^{l \times d}$ represents the input sequence containing attribute $a^t_{\mu}$, where \( l\) is the sequence length and \( d \) is the dimension. The output of the multi-head attention layer, combined with residual connection and layer normalization, is:
\begin{equation}
X'_{a^t_{\mu}} = X_{a^t_{\mu}} + LN(f_{Attn}(X_{a^t_{\mu}} \mid \theta)),
\end{equation}
where $f_{Attn}(\cdot)$ is the multi-head attention layer and $LN(\cdot)$ refers to layer normalization. Each LoRA module $\Delta W_i$ consists of a pair of low-rank matrices \( A_i \in \mathbb{R}^{d_{in} \times r} \) and \( B_i \in \mathbb{R}^{r \times d_{out}} \), where $r \ll min(d_{in},d_{out})$ is the adaptation rank. The LoRA transformation for \(i\)-th module is defined as:
\begin{equation}
L_i(X'_{a^t_{\mu}}) = X'_{a^t_{\mu}} A_i B_i,
\end{equation}
where $1 \leq i \leq n$. $A_i$ and $B_i$ project the input $X'_{a^t_{\mu}}$ to a lower-dimensional space and then back to the original dimension, efficiently capturing task-specific characteristics.

To determine the contribution of each LoRA module and dynamically adjust model parameters, we introduce a gating function $G$, which takes an aspect identifier $D_{A_t}$ as input and outputs a weight vector $\omega \in \mathbb{R}^n$. The gating function is implemented as an embedding layer, a linear layer, and a softmax layer. The output weight is:
\begin{equation}
\omega_i =  \frac{\exp(G(D_A)_i)}{\sum_{j=1}^{n} \exp(G(D_A)_j)},
\end{equation}
where $\omega_{i}$ denotes the weight of the $i$-th LoRA, and the softmax function normalizes the weights. Thus, the output of the feed-forward neural network is:
\begin{equation}
O_{a^t_{\mu}} = f_{FFN}(X'_{a^t_{\mu}}\mid\theta) + \frac{\alpha}{r} \sum_{i=1}^{n} \omega_{i} L_i(X'_{a^t_{\mu}}),
\end{equation}
where $f_{FFN}(\cdot)$ is the feed-forward neural network and $\alpha$ is a constant. Finally, we can get the output of the transformer block, denoted as $H_{a^t_{\mu}}$:
\begin{equation}
H_{a^t_{\mu}} = X'_{a^t_{\mu}} + LN(O_{a^t_{\mu}}).
\end{equation}

We impose some constraints on the output of the last transformer block to reduce the distribution differences between different data sources and balance conflicts between multiple attributes. More details are described in Section \ref{sec:training}.

\subsection{Training Objective}\label{sec:training}
\paragraph{Original Loss: Next Token Prediction $\mathcal{L}_{p}$} In our implement, $\mathcal{L}_{p}$ is computed in the same way as the autoregressive loss of the pre-trained language model, which can align the model output with the target text:
\begin{equation}
\mathcal{L}_{p} = - \sum_{t=1}^{T} \log P_{LM}(y_t \mid x_{<t},D ; \theta),
\end{equation}
where $T$ represents the sequence length, $y_t$ represents the target token at time step $t$ and $x_{<t}$ represents all input tokens before time step $t$.

\paragraph{Aspect-Adaptive Loss $\mathcal{L}_{ada}$} Multi-aspect controllable text generation aims to generate text with multiple attributes from different aspects. Therefore, in the attribute space, the attribute distribution of multi-aspect will be the intersection area of the attribute distribution in multiple aspects. However, due to the distributional differences across data sources, the distribution of different aspects in the attribute space often exhibits minimal overlap, thereby hindering effective multi-attribute text generation. In order to penalize the discrepancy in the distribution center of different aspects at the top of transformer blocks, we use aspect-adaptive loss \cite{gu-etal-2022-distributional}, defined as:
\begin{equation}
\mathcal{L}_{ada} =\!\!\!\!\!\!\!\sum_{1 \leq t_1 < t_2 \leq |\mathcal{A}|} \left\|  \sum_{i=1}^{|I^{t_1}|} \frac{H^i_{a^{t_1}}}{|I^{t_1}|} -  \sum_{j=1}^{|I^{t_2}|} \frac{H^j_{a^{t_2}}}{|I^{t_2}|} \right\|_2,
\end{equation}
where $\left \| \cdot  \right \| _2$ represents the Euclidean distance. It reduces the differences between aspects and enhances the expression of multi-aspect semantics within the attribute space.

\paragraph{Attribute-Aware Loss $\mathcal{L}_{awa}$} In controllable text generation, we need to control specific attributes within specific aspects. For example, in topic control, we aim to generate text with a specific topic. However, since the topic aspect includes multiple categories, we want the model to distinguish between different attributes. To achieve this, we introduce an attribute-aware loss $\mathcal{L}_{awa}$, which consists of two parts: attribute exclusion loss $\mathcal{L}_{e}$ and attribute gap loss $\mathcal{L}_{g}$. 

We aim for the distributions of different attributes in the attribute space to be as independent as possible (in the field of controllable text generation, we only control one attribute within an aspect at a time, not multiple attributes simultaneously). Therefore, we propose the attribute exclusion loss. Consider any two sets of hidden states $H_{a_{\mu_1}^t}$ and $H_{a_{\mu_2}^t}$ with different attributes in aspect $A_t$, the attribute exclusion loss between them is calculated as follows: 
\begin{gather}
    \mathcal{L}_{e}^t = \!\!\!\!\!\!\!\sum_{1 \le \mu_1 < \mu_2 \le |a_t|} \!\!\!\!\!\!\!\max\left( \gamma - \left \| C_{a_{\mu_1}^t} - C_{a_{\mu_2}^t} \right \|_2, 0 \right), \nonumber\\
C_{a_{\mu_i}^t} = \frac{1}{\raisebox{-0.4ex}{\big|} I_{a_{\mu_i}^t} \raisebox{-0.4ex}{\big|} } \sum_{j=1}^{\raisebox{-0.6ex}{\big|}  I_{a_{\mu_i}^t} \raisebox{-0.6ex}{\big|} } H_{a_{\mu_i}^t}^j,
\end{gather}
where $\gamma$ is a hyperparameter, $C_{a_{\mu_i}^t}$ is the distribution center of attribute $a_{\mu_i}^t$ and $\raisebox{-0.4ex}{\big|} I_{a_{\mu_i}^t} \raisebox{-0.4ex}{\big|}$ is the number of hidden states in attribute $a_{\mu_i}^t$. To reduce the difference of the same attribute in the attribute space, we want the distributions within the same attribute to be as cohesive as possible. Therefore, we use attribute gap loss to constrain the distributions within the same attribute to be closer to the center of that attribute’s distribution, which is defined as follows:
\begin{equation}
\mathcal{L}_{g}^t = \sum_{\mu=1}^{\left| a^t \right | } \sum_{j=1 }^{\raisebox{-0.4ex}{\big|} I_{a_{\mu}^t} \raisebox{-0.4ex}{\big|} } \left \| H_{a_{\mu}^t}^{j}-C_{a_{\mu}^t} \right \|_2.
\end{equation}

So the total attribute-aware loss of all aspects is:
\begin{equation}
\mathcal{L}_{awa} = \sum_{t=1}^{\left | \mathcal{A} \right |}{\mathcal{L}_{e}^t + \mathcal{L}_{g}^t}.
\end{equation}

Through attribute awareness, the model can distinguish differences between attributes and search within specific attribute distributions to achieve precise controllable text generation. As shown in Appendix \ref{app:Visualizing attribute distributions}, we visualize the distributions about sentiment and topic aspect in the attribute space, where it is evident that different attribute distributions are as independent as possible, while the distributions of the same attribute are as cohesive as possible.

Our total loss function can be represented as:
\begin{equation}
\mathcal{L} = w_1\mathcal{L}_p + w_2\mathcal{L}_{ada} + w_3\mathcal{L}_{awa}.
\end{equation}

We freeze the parameters of the pre-trained model and only train multiple LoRA modules and the gating function. We obtain significant experimental results, which are detailed in Section \ref{sec:main results}.

\begin{table*}[h!]
\centering
\resizebox{0.9\linewidth}{!}{
\begin{tabular}{l|c|cccccc}
\toprule
\textbf{Model} & \textbf{Average} & \textbf{Sent.} & \textbf{Topic} & \textbf{Multi} & \textbf{Length} & \textbf{Keyword}  & \textbf{Detox.}\\
\midrule
GPT-4 (0613) & \textbf{84.82} &\textbf{91.6} &\textbf{93.5} &\textbf{70.2} &73.8 &\textbf{86.2} &\textbf{93.60}\\
GPT-4o (0513) & 83.10 &91.0 &89.7 &67.6 &\textbf{75.6} &82.0 &92.67\\
\midrule
Qwen2-7B-Instruct &73.16  &85.9	 &91.2	&62.4	&50.1	&58.5	&90.87 \\
\hspace*{4.5em} + LoRA &80.62 &91.0	&\textbf{93.6}	&73.6	&62.6	&72.1	&90.86 \\
\hspace*{4.5em} + FFT &81.54 &\textbf{93.7} &93.2 &75.6 &62.3 &73.8 &90.69\\
\hspace*{4.5em} + Ours &\textbf{82.78} &91.8	&92.7	&\textbf{77.2}	&\textbf{63.9}	&\textbf{79.4}	&\textbf{91.65}\\
\midrule
Qwen2-72B-Instruct &80.30 &87.4	&91.7	&70.5	&62.1	&76.9	&93.21 \\
\hspace*{4.5em} + LoRA &83.85 &92.3 &93.3 &75.7 &68.7 &82.2 &90.92\\
\hspace*{4.5em} + FFT &84.78 &92.4 &\textbf{93.5} &76.4 &70.4 &\textbf{85.2} &90.80\\
\hspace*{4.5em} + Ours &\textbf{85.32} &\textbf{93.1} &92.8 &\textbf{79.4} &\textbf{71.7} &83.4 &\textbf{91.49}\\
\midrule
Llama-3.1-8B-Instruct &75.66  &88.5	&89.9	&64.6 &55.9	&68.5	&86.55 \\
\hspace*{4.5em} + LoRA &79.48 &90.7 &93.9 &71.9 &58.9 &69.9 &91.55\\
\hspace*{4.5em} + FFT &79.47 &\textbf{91.8} &\textbf{94.1} &71.7 &57.2 &71.2 &90.84\\
\hspace*{4.5em} + Ours &\textbf{80.19} &91.3 &93.1 &\textbf{72.2} &\textbf{61.4} &\textbf{71.4} &\textbf{91.73}\\
\midrule
Llama-3.1-70B-Instruct  &81.90 &89.0 &90.9 &70.2 &71.6 &80.6 &89.08\\
\hspace*{4.5em} + LoRA &84.14 &92.4 &\textbf{93.2} &\textbf{74.8} &69.9 &\textbf{83.9} &90.64\\
\hspace*{4.5em} + FFT  &82.60 &88.4 &92.7 &71.1 &69.3 &83.3 &90.80\\
\hspace*{4.5em} + Ours &\textbf{84.61} &\textbf{92.8} &92.5 &73.2 &\textbf{72.7} &83.6 &\textbf{92.86}\\
\midrule
Gemma-2-9B-Instruct &71.73  &88.4 &84.7 &59.0 &53.9 &71.9 &72.49 \\
\hspace*{4.5em} + LoRA &76.91 &89.1 &91.0 &71.5 &50.7  &77.5 &81.67\\
\hspace*{4.5em} + FFT &78.01 &\textbf{90.8} &92.8 &\textbf{74.7} &50.6 &73.5 &\textbf{85.66}\\
\hspace*{4.5em} + Ours &\textbf{79.12} &89.4 &\textbf{92.9} &73.2 &\textbf{57.5} &\textbf{79.3} &82.42\\
\bottomrule
\end{tabular}
}
\caption{Performance of various model variants on different aspects. The model fine-tuned by our framework has the best average accuracy on CoDI-Eval and is ahead of LoRA and FFT in most aspects. After being fine-tuned with our method, Qwen2-72B-Instruct achieved the best performance, surpassing GPT-4. The \textbf{bold} value indicates the maximum value of each model variant on each aspect.}
\label{tab:compare with FFT and LoRA}
\end{table*}

\section{Experiments and Results}\label{experiments}

\subsection{Experiment Setup}

\paragraph{Benchmark dataset.} We conduct experiments on a publicly available controllable text generation benchmark CoDI-Eval \cite{chen2024benchmarkinglargelanguagemodels}. It is a benchmark designed for multi-dimensional fine-grained controllable text generation tasks, encompassing six controllable generation tasks: sentiment, topic, length, keyword, detoxification, and multi, where the multi-aspect generation task requires simultaneous control over sentiment and topic generation. We use GPT-4 to synthesize training data, with approximately 10,000 samples of training data for each task, and use the evaluation data provided by CoDI-Eval for evaluation. More details are provided in Appendix \ref{appendix:benckmark dataset}. 

\paragraph{Baselines.} The methods in related work section primarily achieve controllable text generation by imposing specific constraints and generating subsequent text based on specific prefixes provided by PPLM \cite{dathathri2019plug}. These methods essentially extend text from a given prefix and cannot perform instruction-following tests. Our model architecture is fundamentally different from their approaches, making direct comparison infeasible. As a result, we compare our framework against the following baselines: (1) \textbf{Base Model}: Original pre-trained model without any task-specific fine-tuning. (2) \textbf{LoRA}: Performing low-rank decomposition on specific weight matrices in the pre-trained model enables efficient parameter fine-tuning. (3) \textbf{Full Fine-tuning}: Updating all model parameters to adapt to the specific task fully. See Appendix \ref{app:hyperparameter} for more details on hyperparameter selection.

\paragraph{Evaluation metrics.} We use the fine-tuned RoBERTa classifiers from CoDI-Eval for sentiment, topic, and multi-aspect evaluations, applying rules defined in CoDI-Eval for the length and keyword tasks to conduct the evaluations. We measure the detoxification aspect with the Google Perspective API. Accuracy was used as the evaluation metric for each task, and the model's final performance was calculated as the average accuracy across the six tasks. Detailed information about the classifiers and rules can be found in the Appendix \ref{appendix:Evaluation metrics}.

\subsection{Main results}\label{sec:main results}

Table \ref{tab:compare with FFT and LoRA} shows our method's performance compared to three baselines and the GPT series. Our method consistently achieves higher average accuracy than LoRA and FFT across multiple aspects, reaching optimal results in most of them. For Qwen2-7B-Instruct and Llama-3.1-8B-Instruct, we achieve the best scores in average accuracy, multi-aspect, length, keyword, and detoxification, with up to 1.24\% and 0.72\% improvement over FFT. For Llama-3.1-70B-Instruct and Gemma-2-9B-Instruct, we outperform LoRA and FFT in three aspects and achieve higher average accuracy. Additionally, we outperform GPT-4 for Qwen2-72B-Instruct and achieve top performance on various tasks across all models, demonstrating its superior effectiveness on CoDI-Eval. In Appendix \ref{app:case study}, we present some cases, including attribute combinations provided in CoDI-Eval and unseen attribute combinations, demonstrating the superiority and generalizability of our approach.

\begin{table*}[h!]
\centering
\resizebox{0.91\linewidth}{!}{
\begin{tabular}{l|l|c|cccccc}
\toprule
\textbf{Data Discrepancy} & \textbf{Method} & \textbf{Average} & \textbf{Sent.} & \textbf{Topic} & \textbf{Multi} & \textbf{Length} & \textbf{Keyword}  & \textbf{Detox.}\\
\midrule
\multirow{3}{*}{Sent./Keyword/Multi}  
& LoRA &79.31 &90.9	&91.5	&72.4	&63.8	&66.5	&90.78\\
& FFT  &79.08 &\textbf{92.9} &\textbf{94.1} &\textbf{73.8} &62.3 &60.7 &90.69\\
& Ours &\textbf{80.50} &92.8	&93.2	&72.2	&\textbf{64.8}	&\textbf{68.7}	&\textbf{91.32}\\
\midrule
\multirow{3}{*}{Sent./Length/Keyword} 
& LoRA & 78.15 & \textbf{93.8} & 93.1 & 74.8 & \textbf{50.4} & 66.6 & 90.24 \\
& FFT   &78.11 &92.9 &93.1 &76.5 &47.2 &68.4 &90.56\\
& Ours &\textbf{78.72} &92.4	&\textbf{93.4}	&\textbf{76.5}	&48.6	&\textbf{70.3}	&\textbf{91.14} \\
\midrule
\multirow{3}{*}{Multi/Keyword/Detox.} 
& LoRA  &79.14 &90.8 &92.1 &72.9 &61.9 &66.6 &90.58\\ 
& FFT  &79.24 &\textbf{93.1} &\textbf{93.8} &72.4 &61.0 &64.8 &90.39 \\
& Ours  &\textbf{80.59} &92.0 &93.0 &\textbf{73.9} &\textbf{63.7} &\textbf{70.2} &\textbf{90.73}\\
\bottomrule
\end{tabular}
}
\caption{Performance of Qwen2-7B-Instruct on different metrics under various data discrepancy settings. ‘Data Discrepancy’ means that, for each specific aspect, only 1,000 samples are selected for training. For example, ‘Sent./Keyword/Multi’ indicates that 1,000 samples are randomly selected from each Sentiment, Keyword, and Multi aspects. In contrast, the Topic, Length, and Detoxification aspects use complete samples for training.}
\label{tab:data discrepancy}
\end{table*}

\begin{figure}[t]
    \centering
    \includegraphics[scale=0.32]{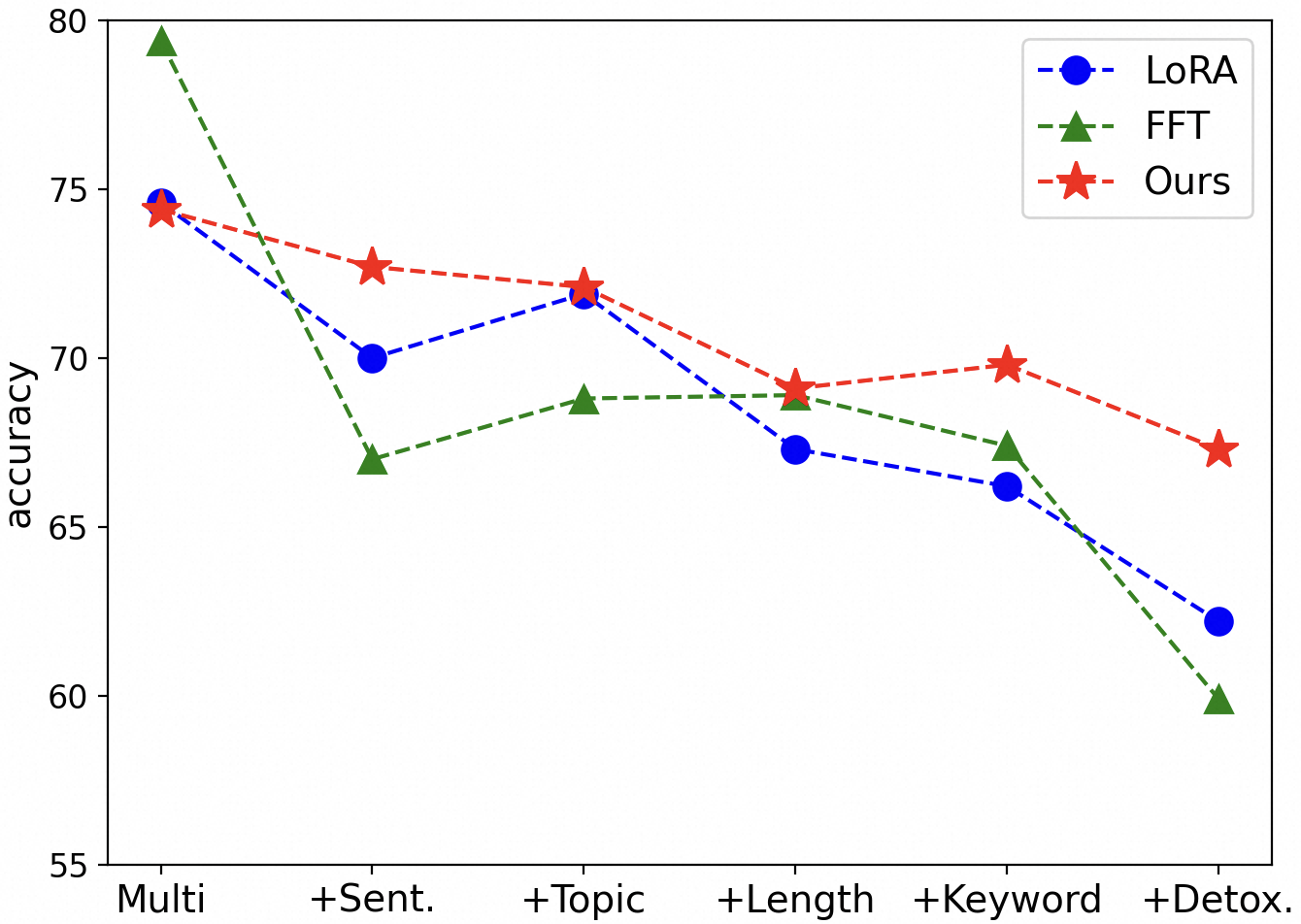}
    \caption{Comparison of LoRA, FFT and our method in alleviating knowledge forgetting. With the injection of new knowledge, we measure the performance of these three methods on multi-aspect task. The performance of our method decreases the least, proving that our method is more robust to knowledge forgetting.}
    \label{fig:knowledge}
\end{figure}

Moreover, we evaluate the performance of LoRA, FFT and our method under varying training data discrepancies. Specifically, we set training data for any three aspects to 1,000 samples each while using the complete training data for the remaining three aspects. The results are presented in Table \ref{tab:data discrepancy}. We observe that, under conditions of imbalanced training data, our method consistently achieves the highest average accuracy and maintains superior performance across most aspects, with at most a 1.42\% significant improvement over existing baselines. This demonstrates the adaptability of our approach to varying data discrepancies.

Alleviating knowledge forgetting is a significant research topic in the field of natural language processing \cite{huang-etal-2024-mitigating,li-etal-2024-revisiting,dou-etal-2024-loramoe}. To evaluate the ability of out method to alleviate knowledge forgetting, we conduct the following experiment: First, we fine-tune Qwen2-7B-Instruct on the multi-aspect task using LoRA, FFT and our method respectively, recording their performance. Subsequently, we iteratively fine-tune each model on a new aspect and test their performance on the original multi-aspect task. The results, shown in Figure \ref{fig:knowledge}, indicate that as more knowledge is injected, the performance of LoRA and FFT significantly deteriorates on the multi-aspect task, whereas our method experiences only a slight decrease. This demonstrates that our method is more robust in terms of knowledge forgetting.

\begin{figure}[t]
    \centering
    \includegraphics[scale=0.175]{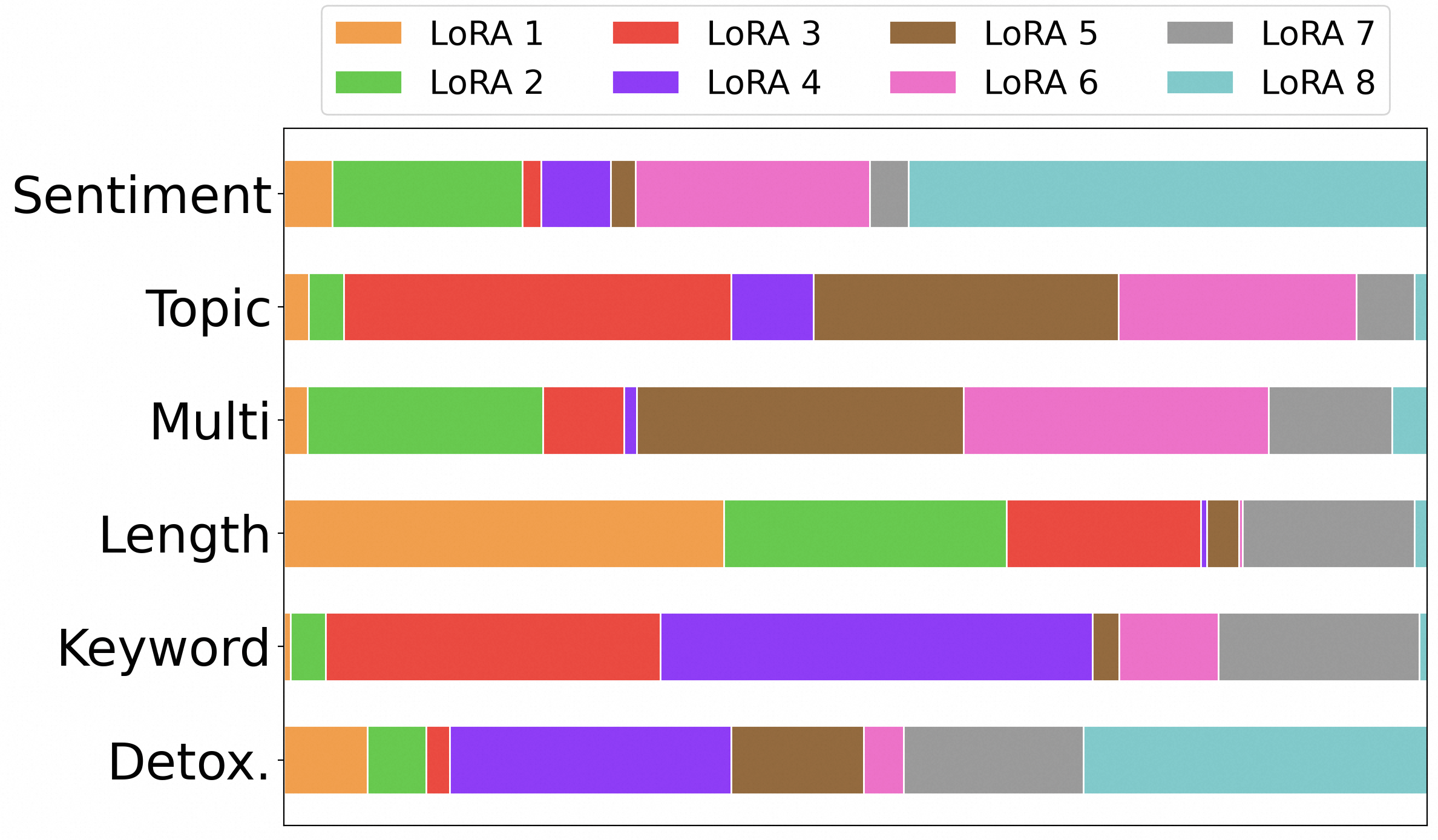}
    \caption{The visualization of LoRA weights for various aspects based on Qwen2-7B-Instruct.}
    \label{fig:qwen2-7b-weight}
\end{figure}

\section{Analysis}\label{sec:analysis}
\paragraph{Visualizing the LoRA weights.} In Figure \ref{fig:qwen2-7b-weight}, we visualize the LoRA weights across six aspects of CoDI-Eval. The bar lengths in different colors represent each LoRA's weight. Our findings indicate that the contributions of LoRAs vary significantly across different aspects, highlighting their specialized roles in distinct control domains. Additionally, within each control aspect, the variation in contributions among different LoRAs is also quite pronounced, demonstrating that our framework effectively trains LoRA to specialize in different control domains and efficiently integrates their expertise, learning the differences between aspects to achieve precise controllable text generation, which neither the original LoRA nor FFT can accomplish.

\paragraph{Compare with Independent LoRA per Aspect.} To further evaluate the effectiveness of our framework, we compare it against a variant of our framework where each aspect is trained with a separate LoRA module, rather than integrating multiple aspects within a unified model. Specifically, we train independent LoRA modules for each of the six aspects in CoDI-Eval and evaluate their performance on the benchmark. Table \ref{tab:framework variant} presents the comparison results. The results demonstrate that our unified LoRA framework consistently outperforms the independent LoRA approach, particularly in multi-aspect controllability, where our framework achieves a 3.0\% improvement. This suggests that dynamically adjusting LoRA combinations through our gating mechanism enables better multi-attribute interaction while maintaining control precision.

\begin{table*}[t]
\centering
\resizebox{0.9\linewidth}{!}{
\begin{tabular}{l|c|cccccc}
\toprule
\textbf{Model} & \textbf{Average} & \textbf{Sent.} & \textbf{Topic} & \textbf{Multi} & \textbf{Length} & \textbf{Keyword}  & \textbf{Detox.}\\
\midrule
Independent LoRA per Aspect &81.22 &91.3	&\textbf{93.0}	&74.2	&62.3	&75.8	&90.72\\
Our framework  &\textbf{82.78} &\textbf{91.8}	&92.7	&\textbf{77.2}	&\textbf{63.9}	&\textbf{79.4}	&\textbf{91.65}\\
\bottomrule
\end{tabular}
}
\caption{Comparison of our implementation with independent LoRA per aspect.}
\label{tab:framework variant}
\end{table*}

\begin{table*}[h!]
\centering
\resizebox{0.9\linewidth}{!}{
\begin{tabular}{l|c|cccccc}
\toprule
\textbf{Routing Strategies} & \textbf{Average} & \textbf{Sent.} & \textbf{Topic} & \textbf{Multi} & \textbf{Length} & \textbf{Keyword}  & \textbf{Detox.}\\
\midrule
All LoRA Modules &\textbf{82.78} &\textbf{91.8}	&\textbf{92.7}	&\textbf{77.2}	&\textbf{63.9}	&79.4	&91.65\\
Top-2 LoRA Modules  &81.94 &88.5 &92.4 &74.9 &62.2 &\textbf{80.2} &\textbf{93.52}\\
\bottomrule
\end{tabular}
}
\caption{Compare the effect of each token passing through all LoRA modules and the Top-2 LoRA modules.}
\label{tab:compare different routing strategies}
\end{table*}

\begin{table*}[h!]
	\centering
    \resizebox{0.8\linewidth}{!}{
	\begin{tabular}{ccccc}
		\toprule
		\textbf{Model}  & \textbf{w/o $\mathcal{L}_{ada}$ and $\mathcal{L}_{awa}$} &\textbf{w/o  $\mathcal{L}_{awa}$} &\textbf{w/o  $\mathcal{L}_{ada}$} & \textbf{Ours} \\
		\midrule
        Qwen2-7B-Instruct       & 81.83        & 82.11   & 82.42         & \textbf{82.78}\\
        Gemma-2-9B-Instruct       & 77.48          & 77.55        &77.82
        & \textbf{79.12} \\
		\bottomrule
	\end{tabular}
    }
	\caption{Evaluation average results for loss ablation on Qwen2-7B-Instruct and Gemma-2-9B-Instruct.}
    \label{tab:ablation loss}
\end{table*}

\paragraph{Compare different routing strategies.} In our framework, tokens pass through all LoRA modules. To evaluate the impact of different routing strategies, we compare it with the Top-2 routing strategy and the experimental results are shown in Table \ref{tab:compare different routing strategies}. We found that when the token passes through all LoRA modules, the model performs better and outperforms the Top-2 routing strategy in most aspects. This is because there is synergy between different tasks, and using more LoRA modules allows for capturing more information, thereby achieving better control effects.

\paragraph{Ablation on $\mathcal{L}_{ada}$ and $\mathcal{L}_{awa}$.} We conduct ablation experiments on losses based on Qwen2-7B-Instruct and Gemma-2-9B-Instruct to verify the impact of each loss on the average result, and the results are shown in Table \ref{tab:ablation loss}. In Table \ref{tab:ablation loss}, we compare four different settings of our framework, and "w/o $\mathcal{L}_{ada}$ and $\mathcal{L}_{awa}$" is a variant without aspect adaptation and attribute awareness, which performs the worst. In addition, $\mathcal{L}_{awa}$ has a higher impact on the average result than $\mathcal{L}_{ada}$. With all enhancements included, our framework performs best.

\paragraph{Selection of the Number and Rank of LoRA.} We examine the effect of varying the number and rank of LoRA modules on model performance. As shown in Table \ref{tab:para}, our findings reveal that the optimal performance is achieved when the number of LoRA modules is set to 8, with each LoRA's rank $r$ fixed at 16. Increasing the number of LoRA modules beyond 8 does not lead to further performance gains. Additionally, while increasing the LoRA rank to 32 results in a slight performance improvement, it also doubles the number of training parameters. To balance model efficiency and performance, we select a final rank of 16.

\begin{table}[t]
	\centering
    \resizebox{0.9\linewidth}{!}{
	\label{tab:1}
	\begin{tabular}{cccc}
		\toprule
		\textbf{LoRAs}  & \tabincell{c}{\textbf{LoRA} \\ \textbf{Rank}} & \tabincell{c}{\textbf{Trainable} \\\textbf{Param.}} & \tabincell{c}{\textbf{Average}} \\
		\midrule
        4       & 16            & 2.04\%            & 81.58\\
        8       & 16            & 4.07\%            & 82.78\\
        16       & 16            & 8.14\%            & 82.33\\
        \midrule
        8       & 8            & 2.04\%            & 82.10\\
        8       & 16            & 4.07\%            & 82.78\\
        8       & 32            & 8.14\%            & 82.93\\
		\bottomrule
	\end{tabular}
    }
	\caption{Performance of Our framework
 varies with the number of LoRAs and LoRA rank across all aspects.}
    \label{tab:para}
\end{table}

\section{Conclusion}
We propose a lightweight, adaptive, and attribute-aware framework for multi-aspect controllable text generation, achieving excellent performance on controllable text generation tasks. Our framework integrates multiple LoRA modules and employs a trainable gating function to combine LoRAs for each control aspect optimally. Through comprehensive experiments, we demonstrate that our method outperforms existing LoRA and FFT approaches in controllable text generation tasks, achieving SOTA. Our framework is adaptable to imbalanced data and effectively retains previously learned knowledge.

\section*{Limitations}
Although our method demonstrates promising results, it also has several limitations. (i) CoDI-Eval primarily evaluates instruction-following controllability rather than real-world text generation needs, where both content and attributes may require explicit control. While this makes it suitable for benchmarking attribute control, its applicability to broader contexts remains limited. (ii) Like existing benchmarks, CoDI-Eval relies on fixed attribute sets, limiting generalization to unseen attributes. This is a common challenge in controllable text generation rather than a constraint of our framework. Future work could explore more flexible benchmarks supporting dynamic attribute adaptation. (iii) In high-conflict scenarios, where attributes exhibit significant contradictions (e.g., generating text with simultaneously positive and negative sentiments), the quality of generated text may degrade. While our method alleviates attribute conflicts through the attribute-aware loss function, these mechanisms may not be sufficient to fully resolve severe attribute contradictions, particularly as the number of control attributes increases.

\bibliography{main}

\begin{thebibliography}{26}
\providecommand{\natexlab}[1]{#1}

\bibitem[{Carlsson et~al.(2022)Carlsson, {\"O}hman, Liu, Verlinden, Nivre, and Sahlgren}]{carlsson-etal-2022-fine}
Fredrik Carlsson, Joey {\"O}hman, Fangyu Liu, Severine Verlinden, Joakim Nivre, and Magnus Sahlgren. 2022.
\newblock \href {https://doi.org/10.18653/v1/2022.acl-long.471} {Fine-grained controllable text generation using non-residual prompting}.
\newblock In \emph{Proceedings of the 60th Annual Meeting of the Association for Computational Linguistics (Volume 1: Long Papers)}, pages 6837--6857, Dublin, Ireland. Association for Computational Linguistics.

\bibitem[{Chan et~al.(2021)Chan, Madani, Krause, and Naik}]{chan2021deep}
Alvin Chan, Ali Madani, Ben Krause, and Nikhil Naik. 2021.
\newblock Deep extrapolation for attribute-enhanced generation.
\newblock \emph{Advances in Neural Information Processing Systems}, 34:14084--14096.

\bibitem[{Chen et~al.(2024)Chen, Xu, Wang, Liu, and Mao}]{chen2024benchmarkinglargelanguagemodels}
Yihan Chen, Benfeng Xu, Quan Wang, Yi~Liu, and Zhendong Mao. 2024.
\newblock \href {https://arxiv.org/abs/2401.00690} {Benchmarking large language models on controllable generation under diversified instructions}.
\newblock \emph{Preprint}, arXiv:2401.00690.

\bibitem[{Dathathri et~al.(2019)Dathathri, Madotto, Lan, Hung, Frank, Molino, Yosinski, and Liu}]{dathathri2019plug}
Sumanth Dathathri, Andrea Madotto, Janice Lan, Jane Hung, Eric Frank, Piero Molino, Jason Yosinski, and Rosanne Liu. 2019.
\newblock Plug and play language models: A simple approach to controlled text generation.
\newblock \emph{arXiv preprint arXiv:1912.02164}.

\bibitem[{Dou et~al.(2024)Dou, Zhou, Liu, Gao, Shen, Xiong, Zhou, Wang, Xi, Fan, Pu, Zhu, Zheng, Gui, Zhang, and Huang}]{dou-etal-2024-loramoe}
Shihan Dou, Enyu Zhou, Yan Liu, Songyang Gao, Wei Shen, Limao Xiong, Yuhao Zhou, Xiao Wang, Zhiheng Xi, Xiaoran Fan, Shiliang Pu, Jiang Zhu, Rui Zheng, Tao Gui, Qi~Zhang, and Xuanjing Huang. 2024.
\newblock \href {https://doi.org/10.18653/v1/2024.acl-long.106} {{L}o{RAM}o{E}: Alleviating world knowledge forgetting in large language models via {M}o{E}-style plugin}.
\newblock In \emph{Proceedings of the 62nd Annual Meeting of the Association for Computational Linguistics (Volume 1: Long Papers)}, pages 1932--1945, Bangkok, Thailand. Association for Computational Linguistics.

\bibitem[{Feng et~al.(2023)Feng, Yi, Wang, Lakshmanan, and Xie}]{feng-etal-2023-dunst}
Yuxi Feng, Xiaoyuan Yi, Xiting Wang, Laks Lakshmanan, V.S., and Xing Xie. 2023.
\newblock \href {https://doi.org/10.18653/v1/2023.acl-long.488} {{D}u{NST}: Dual noisy self training for semi-supervised controllable text generation}.
\newblock In \emph{Proceedings of the 61st Annual Meeting of the Association for Computational Linguistics (Volume 1: Long Papers)}, pages 8760--8785, Toronto, Canada. Association for Computational Linguistics.

\bibitem[{Gu et~al.(2022)Gu, Feng, Ma, Zhang, Gong, and Qin}]{gu-etal-2022-distributional}
Yuxuan Gu, Xiaocheng Feng, Sicheng Ma, Lingyuan Zhang, Heng Gong, and Bing Qin. 2022.
\newblock \href {https://doi.org/10.18653/v1/2022.emnlp-main.67} {A distributional lens for multi-aspect controllable text generation}.
\newblock In \emph{Proceedings of the 2022 Conference on Empirical Methods in Natural Language Processing}, pages 1023--1043, Abu Dhabi, United Arab Emirates. Association for Computational Linguistics.

\bibitem[{Gu et~al.(2023)Gu, Feng, Ma, Zhang, Gong, Zhong, and Qin}]{gu-etal-2023-controllable}
Yuxuan Gu, Xiaocheng Feng, Sicheng Ma, Lingyuan Zhang, Heng Gong, Weihong Zhong, and Bing Qin. 2023.
\newblock \href {https://doi.org/10.18653/v1/2023.acl-long.704} {Controllable text generation via probability density estimation in the latent space}.
\newblock In \emph{Proceedings of the 61st Annual Meeting of the Association for Computational Linguistics (Volume 1: Long Papers)}, pages 12590--12616, Toronto, Canada. Association for Computational Linguistics.

\bibitem[{Hu et~al.(2022)Hu, Shen, Wallis, Allen-Zhu, Li, Wang, Wang, and Chen}]{hu2022lora}
Edward~J Hu, Yelong Shen, Phillip Wallis, Zeyuan Allen-Zhu, Yuanzhi Li, Shean Wang, Lu~Wang, and Weizhu Chen. 2022.
\newblock \href {https://openreview.net/forum?id=nZeVKeeFYf9} {Lo{RA}: Low-rank adaptation of large language models}.
\newblock In \emph{International Conference on Learning Representations}.

\bibitem[{Huang et~al.(2024)Huang, Cui, Wang, Yang, Liao, Song, Yao, and Su}]{huang-etal-2024-mitigating}
Jianheng Huang, Leyang Cui, Ante Wang, Chengyi Yang, Xinting Liao, Linfeng Song, Junfeng Yao, and Jinsong Su. 2024.
\newblock \href {https://doi.org/10.18653/v1/2024.acl-long.77} {Mitigating catastrophic forgetting in large language models with self-synthesized rehearsal}.
\newblock In \emph{Proceedings of the 62nd Annual Meeting of the Association for Computational Linguistics (Volume 1: Long Papers)}, pages 1416--1428, Bangkok, Thailand. Association for Computational Linguistics.

\bibitem[{Keskar et~al.(2019)Keskar, McCann, Varshney, Xiong, and Socher}]{keskarCTRL2019}
Nitish~Shirish Keskar, Bryan McCann, Lav Varshney, Caiming Xiong, and Richard Socher. 2019.
\newblock {CTRL - A Conditional Transformer Language Model for Controllable Generation}.
\newblock \emph{arXiv preprint arXiv:1909.05858}.

\bibitem[{Krause et~al.(2021)Krause, Gotmare, McCann, Keskar, Joty, Socher, and Rajani}]{krause-etal-2021-gedi-generative}
Ben Krause, Akhilesh~Deepak Gotmare, Bryan McCann, Nitish~Shirish Keskar, Shafiq Joty, Richard Socher, and Nazneen~Fatema Rajani. 2021.
\newblock \href {https://doi.org/10.18653/v1/2021.findings-emnlp.424} {{G}e{D}i: Generative discriminator guided sequence generation}.
\newblock In \emph{Findings of the Association for Computational Linguistics: EMNLP 2021}, pages 4929--4952, Punta Cana, Dominican Republic. Association for Computational Linguistics.

\bibitem[{Kumar et~al.(2023)Kumar, Koorehdavoudi, Moshtaghi, Misra, Chadha, and Ferrara}]{kumar-etal-2023-controlled}
Vaibhav Kumar, Hana Koorehdavoudi, Masud Moshtaghi, Amita Misra, Ankit Chadha, and Emilio Ferrara. 2023.
\newblock \href {https://doi.org/10.18653/v1/2023.findings-acl.602} {Controlled text generation with hidden representation transformations}.
\newblock In \emph{Findings of the Association for Computational Linguistics: ACL 2023}, pages 9440--9455, Toronto, Canada. Association for Computational Linguistics.

\bibitem[{Li et~al.(2024)Li, Ding, Fang, and Tao}]{li-etal-2024-revisiting}
Hongyu Li, Liang Ding, Meng Fang, and Dacheng Tao. 2024.
\newblock \href {https://aclanthology.org/2024.findings-emnlp.249} {Revisiting catastrophic forgetting in large language model tuning}.
\newblock In \emph{Findings of the Association for Computational Linguistics: EMNLP 2024}, pages 4297--4308, Miami, Florida, USA. Association for Computational Linguistics.

\bibitem[{Li and Liang(2021)}]{li-liang-2021-prefix}
Xiang~Lisa Li and Percy Liang. 2021.
\newblock \href {https://doi.org/10.18653/v1/2021.acl-long.353} {Prefix-tuning: Optimizing continuous prompts for generation}.
\newblock In \emph{Proceedings of the 59th Annual Meeting of the Association for Computational Linguistics and the 11th International Joint Conference on Natural Language Processing (Volume 1: Long Papers)}, pages 4582--4597, Online. Association for Computational Linguistics.

\bibitem[{Li et~al.(2022)Li, Thickstun, Gulrajani, Liang, and Hashimoto}]{Li-2022-DiffusionLM}
Xiang~Lisa Li, John Thickstun, Ishaan Gulrajani, Percy Liang, and Tatsunori Hashimoto. 2022.
\newblock Diffusion-lm improves controllable text generation.
\newblock \emph{ArXiv}, abs/2205.14217.

\bibitem[{Liu et~al.(2024)Liu, Liu, Zhu, and Hu}]{liu-etal-2024-multi}
Yi~Liu, Xiangyu Liu, Xiangrong Zhu, and Wei Hu. 2024.
\newblock \href {https://doi.org/10.18653/v1/2024.acl-long.500} {Multi-aspect controllable text generation with disentangled counterfactual augmentation}.
\newblock In \emph{Proceedings of the 62nd Annual Meeting of the Association for Computational Linguistics (Volume 1: Long Papers)}, pages 9231--9253, Bangkok, Thailand. Association for Computational Linguistics.

\bibitem[{Lu et~al.(2023)Lu, Wei, Qu, Mao, Chen, and Chen}]{lu-etal-2023-miracle}
Zhenyi Lu, Wei Wei, Xiaoye Qu, Xian-Ling Mao, Dangyang Chen, and Jixiong Chen. 2023.
\newblock \href {https://doi.org/10.18653/v1/2023.findings-emnlp.395} {Miracle: Towards personalized dialogue generation with latent-space multiple personal attribute control}.
\newblock In \emph{Findings of the Association for Computational Linguistics: EMNLP 2023}, pages 5933--5957, Singapore. Association for Computational Linguistics.

\bibitem[{Shuttleworth et~al.(2024)Shuttleworth, Andreas, Torralba, and Sharma}]{shuttleworth2024lora}
Reece Shuttleworth, Jacob Andreas, Antonio Torralba, and Pratyusha Sharma. 2024.
\newblock Lo{RA} vs {F}ull {F}ine-tuning: An illusion of equivalence.
\newblock \emph{arXiv preprint arXiv:2410.21228}.

\bibitem[{Yang et~al.(2023)Yang, Liu, Lei, Yang, Xue, Chen, and Xie}]{yang-etal-2023-tailor}
Kexin Yang, Dayiheng Liu, Wenqiang Lei, Baosong Yang, Mingfeng Xue, Boxing Chen, and Jun Xie. 2023.
\newblock \href {https://doi.org/10.18653/v1/2023.acl-long.25} {Tailor: A soft-prompt-based approach to attribute-based controlled text generation}.
\newblock In \emph{Proceedings of the 61st Annual Meeting of the Association for Computational Linguistics (Volume 1: Long Papers)}, pages 410--427, Toronto, Canada. Association for Computational Linguistics.

\bibitem[{Zeldes et~al.(2020)Zeldes, Padnos, Sharir, and Peleg}]{zeldes2020technicalreportauxiliarytuning}
Yoel Zeldes, Dan Padnos, Or~Sharir, and Barak Peleg. 2020.
\newblock \href {https://arxiv.org/abs/2006.16823} {Technical report: Auxiliary tuning and its application to conditional text generation}.
\newblock \emph{Preprint}, arXiv:2006.16823.

\bibitem[{Zhang and Song(2022)}]{zhang-song-2022-discup}
Hanqing Zhang and Dawei Song. 2022.
\newblock \href {https://doi.org/10.18653/v1/2022.emnlp-main.223} {{D}is{C}up: Discriminator cooperative unlikelihood prompt-tuning for controllable text generation}.
\newblock In \emph{Proceedings of the 2022 Conference on Empirical Methods in Natural Language Processing}, pages 3392--3406, Abu Dhabi, United Arab Emirates. Association for Computational Linguistics.

\bibitem[{Zheng et~al.(2023)Zheng, Ke, Zhang, and Huang}]{zheng-etal-2023-click}
Chujie Zheng, Pei Ke, Zheng Zhang, and Minlie Huang. 2023.
\newblock \href {https://doi.org/10.18653/v1/2023.findings-acl.65} {Click: Controllable text generation with sequence likelihood contrastive learning}.
\newblock In \emph{Findings of the Association for Computational Linguistics: ACL 2023}, pages 1022--1040, Toronto, Canada. Association for Computational Linguistics.

\bibitem[{Zhong et~al.(2023)Zhong, Wang, Han, Zhang, and Mao}]{zhong-etal-2023-air}
Tianqi Zhong, Quan Wang, Jingxuan Han, Yongdong Zhang, and Zhendong Mao. 2023.
\newblock \href {https://doi.org/10.18653/v1/2023.emnlp-main.512} {Air-decoding: Attribute distribution reconstruction for decoding-time controllable text generation}.
\newblock In \emph{Proceedings of the 2023 Conference on Empirical Methods in Natural Language Processing}, pages 8233--8248, Singapore. Association for Computational Linguistics.

\bibitem[{Zhou et~al.(2023)Zhou, Jiang, Wilcox, Cotterell, and Sachan}]{pmlr-v202-zhou23g}
Wangchunshu Zhou, Yuchen~Eleanor Jiang, Ethan Wilcox, Ryan Cotterell, and Mrinmaya Sachan. 2023.
\newblock \href {https://proceedings.mlr.press/v202/zhou23g.html} {Controlled text generation with natural language instructions}.
\newblock In \emph{Proceedings of the 40th International Conference on Machine Learning}, volume 202 of \emph{Proceedings of Machine Learning Research}, pages 42602--42613. PMLR.

\bibitem[{Ziegler et~al.(2020)Ziegler, Stiennon, Wu, Brown, Radford, Amodei, Christiano, and Irving}]{ziegler2020finetuninglanguagemodelshuman}
Daniel~M. Ziegler, Nisan Stiennon, Jeffrey Wu, Tom~B. Brown, Alec Radford, Dario Amodei, Paul Christiano, and Geoffrey Irving. 2020.
\newblock \href {https://arxiv.org/abs/1909.08593} {Fine-tuning language models from human preferences}.
\newblock \emph{Preprint}, arXiv:1909.08593.

\end{thebibliography}

\appendix

\section{Details of Hyperparameter Selection}
\label{app:hyperparameter}
In this section, we describe the hyperparameters used for our framework. In our approach, multiple LoRA modules are applied to all linear layers of the pre-trained model, with the number of LoRA modules $n$ set to 8, the rank $r$ set to 16, the scaling factor $\alpha$ set to 32, and dropout set to 0.1. The gating function is implemented using an embedding layer and a linear layer. The input dimension of the embedding layer is the number of aspects, which is 6, and the output dimension is 64. The input dimension of the linear layer is 64, and the output dimension is 8.

We use 8 NVIDIA A100 80GB GPUs during training and employ \texttt{bfloat16} precision to improve training efficiency. The batch size is set to 64. The scaling factors for $\mathcal{L}_{p}$, $\mathcal{L}_{ada}$, and $\mathcal{L}_{awa}$ are 0.7, 0.2, and 0.1, respectively and hyperparameter $\gamma$ in $\mathcal{L}_{awa}$ is set to 1. The optimizer used in our experiments is AdamW. For models with a scale below 10B, we set the learning rate to $2 \times 10^{-4}$ to ensure efficient convergence while maintaining stability during training. For larger models, specifically the 70B and 72B scale, we adjust the learning rate to $1 \times 10^{-5}$ to accommodate the increased model complexity and prevent issues such as gradient instability or overfitting. For models with fewer than 10B parameters, the number of training epochs is 9, while for models with 70B and 72B parameters, the number of epochs is 3.

During inference, we use specific sampling hyperparameters: $max\_new\_tokens$ is set to 512, $top\_p$ is set to 0.7, and $temperature$ is set to 0.95.

\begin{figure}[t]
    \centering
    \includegraphics[scale=0.175]{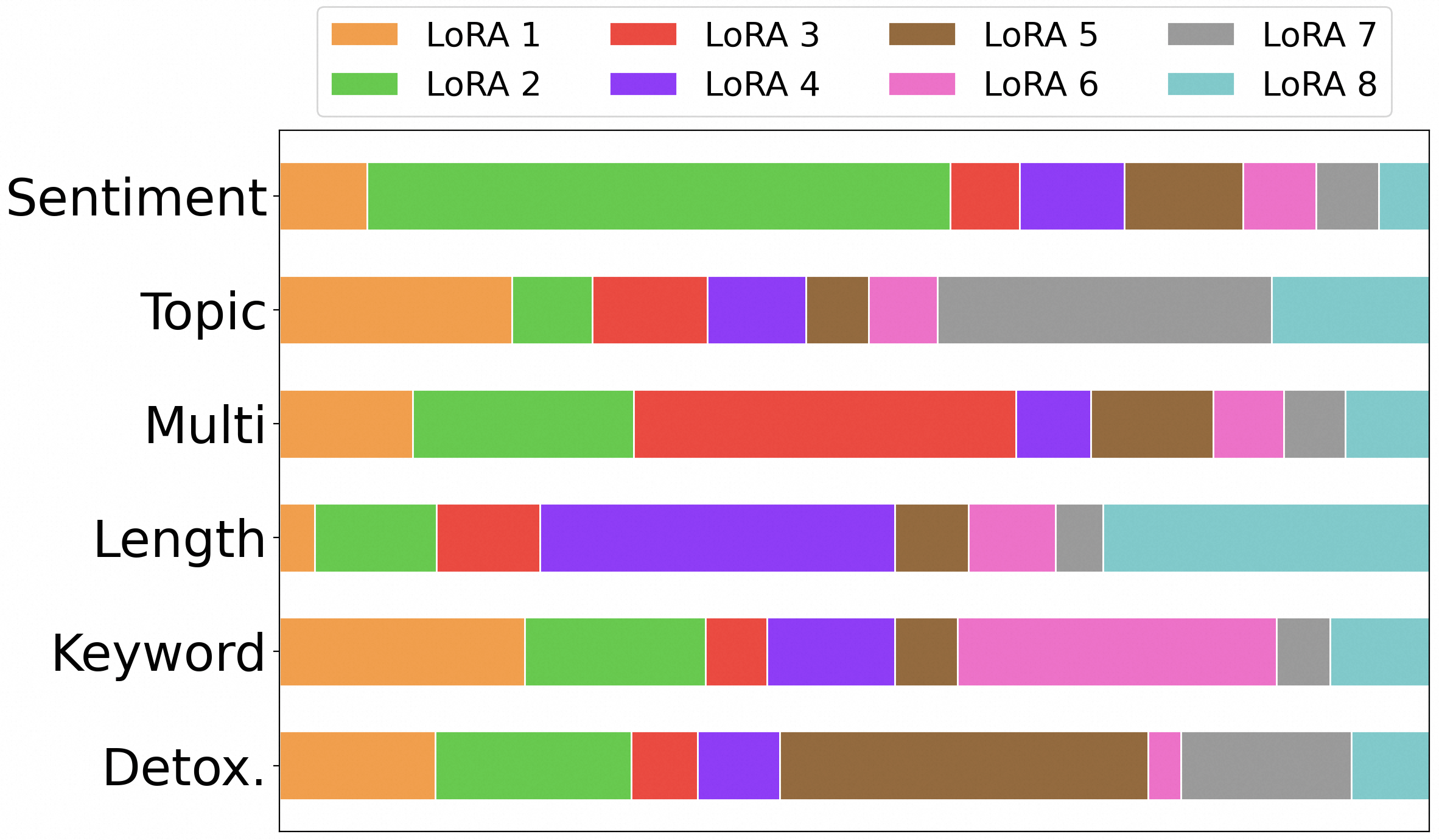}
    \caption{The visualization of LoRA weights for various aspects based on Qwen2-72B-Instruct.}
    \label{fig:qwen2-72b-weight}
\end{figure}

\begin{figure}[t]
    \centering
    \includegraphics[scale=0.175]{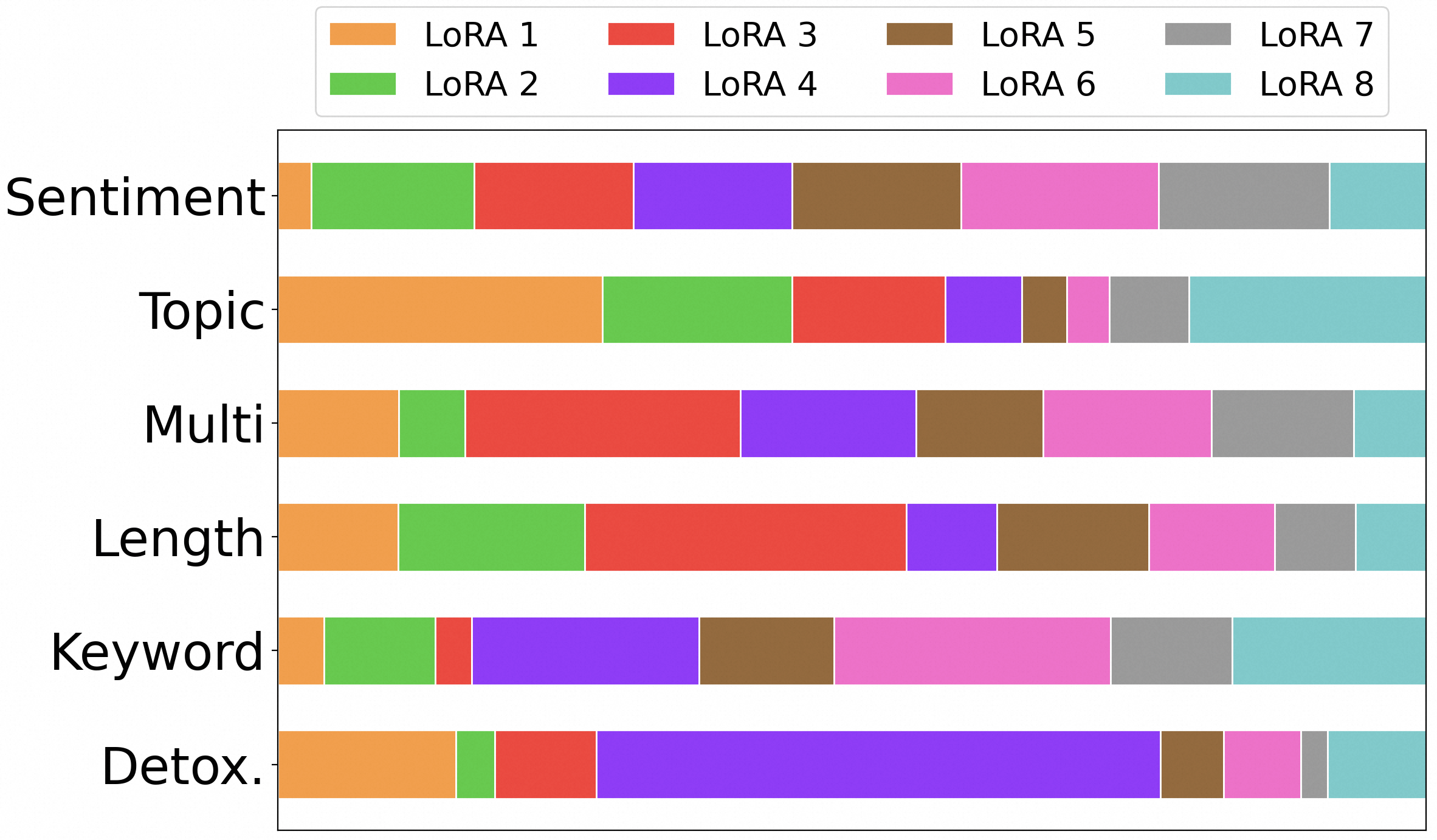}
    \caption{The visualization of LoRA weights for various aspects based on Llama-3.1-8B-Instruct.}
    \label{fig:llama3.1-8b-weight}
\end{figure}

\begin{figure}[t]
    \centering
    \includegraphics[scale=0.175]{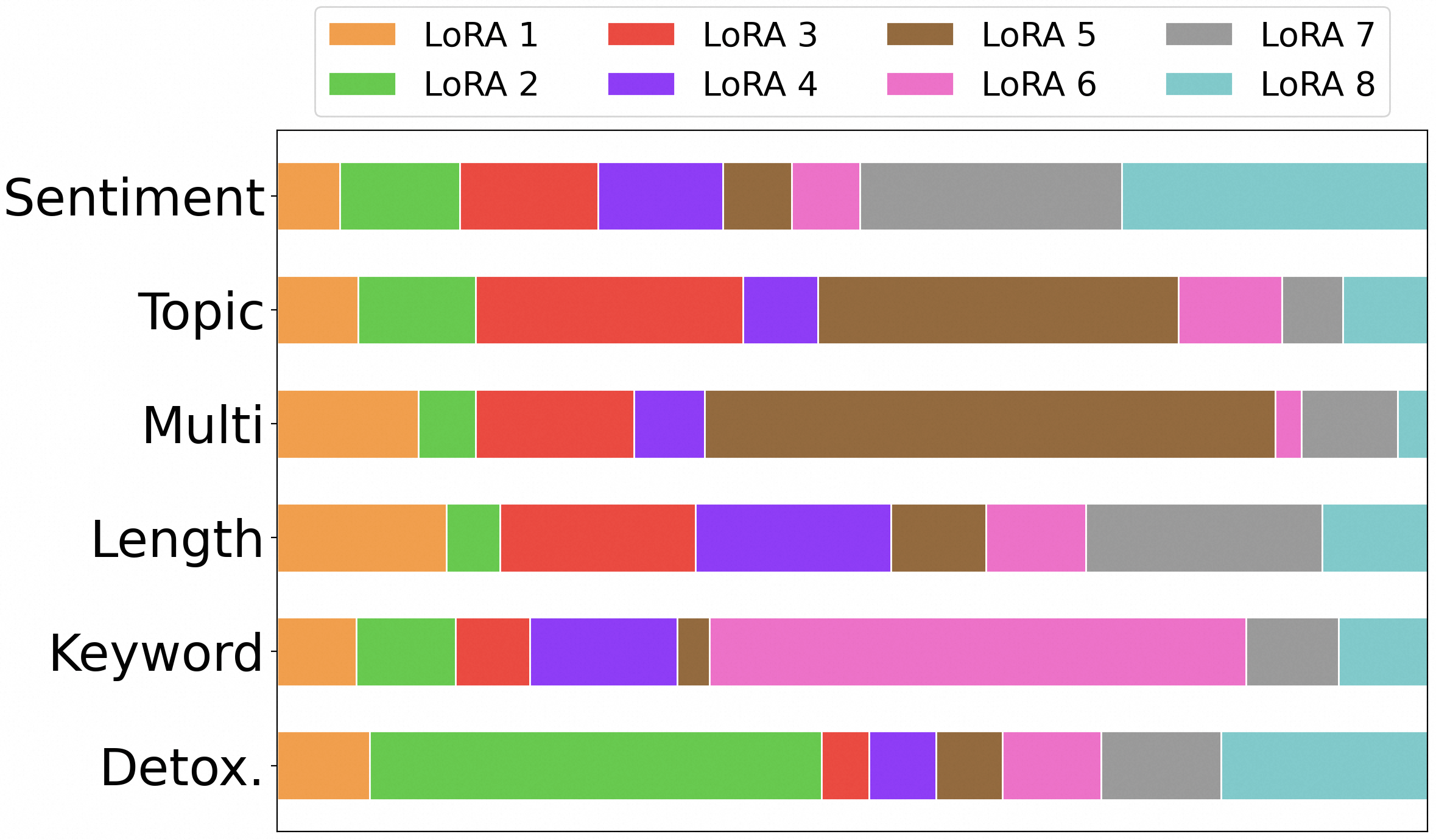}
    \caption{The visualization of LoRA weights for various aspects based on Llama-3.1-70B-Instruct.}
    \label{fig:llama3.1-70b-weight}
\end{figure}

\begin{figure}[t]
    \centering
    \includegraphics[scale=0.175]{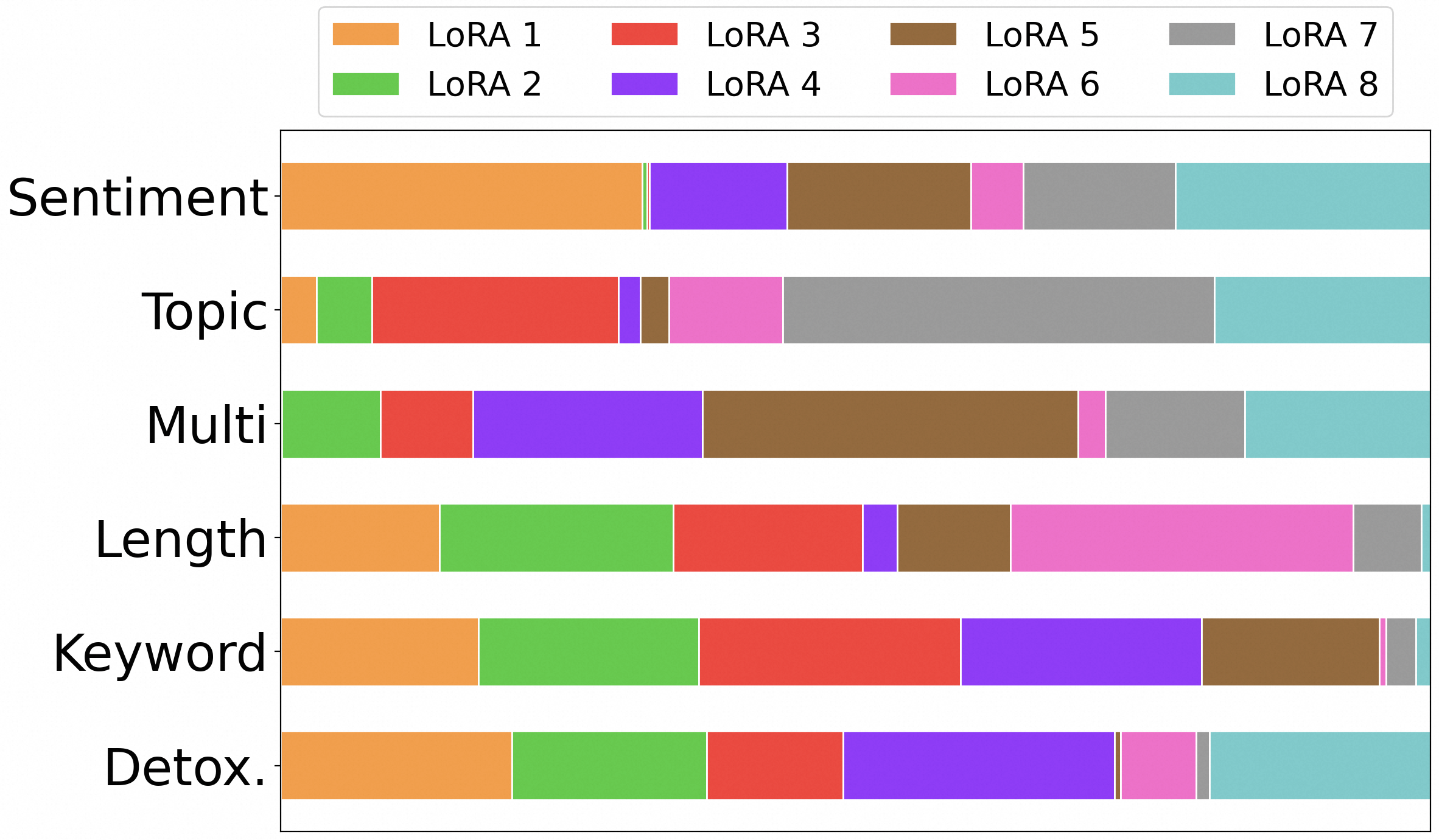}
    \caption{The visualization of LoRA weights for various aspects based on Gemma-2-9B-Instruct.}
    \label{fig:gemma2-weight}
\end{figure}

\begin{table*}[ht]
\centering
\resizebox{0.9\linewidth}{!}{
\begin{tabular}{c|c|cccccc}
\toprule
\textbf{Input to the gating function} & \textbf{Average} & \textbf{Sent.} & \textbf{Topic} & \textbf{Multi} & \textbf{Length} & \textbf{Keyword}  & \textbf{Detox.}\\
\midrule
Aspect Identifier &\textbf{82.78} &91.8	&\textbf{92.7}	&\textbf{77.2}	&\textbf{63.9}	&79.4	&\textbf{91.65}\\
Hidden States &81.23 &\textbf{92.2} &92.5 &72.4 &59.2 &\textbf{80.2} &90.90 \\
\bottomrule
\end{tabular}
}
\caption{Compare the impact of different inputs to the gating function on model performance.}
\label{tab:input to the gating function}
\end{table*}

\section{Details of Experiment Setup}\label{sec:Details of Experiment Setup}
\subsection{Benchmark Dataset}\label{appendix:benckmark dataset}
We use CoDI-Eval as the benchmark for our experiments. CoDI-Eval encompasses six controllable text generation tasks: sentiment, topic, multi, length, keyword, and detoxification. The sentiment and topic control tasks require generating text with specified sentiments and topics. The multi-aspect control task involves fulfilling both sentiment and topic requirements simultaneously. For the keyword control task, the generated text must include specified keywords, while the length control task demands controlling the length of the output. Lastly, the detoxification task ensures that the generated text is free of toxic content. Since this benchmark does not provide training data, we synthesized training data using GPT-4: the sentiment control task contains 8,798 samples, the detoxification control task contains 11,753 samples, the keyword control task contains 8,294 samples, the length control task contains 19,238 samples, the topic control task contains 10,599 samples, and the multi-control task contains 12,678 samples, totaling 71,320 training samples. The test data provided by CoDI-Eval, with the detoxification control task containing 4,060 test samples and the other five control tasks, each containing 1,000 test samples. We labeled the training and test data with aspect identifiers numbered 0 to 5, respectively. To ensure reproducibility and facilitate further research, we will release all training data, evaluation scripts, and the fine-tuned model checkpoints. The dataset includes synthesized training samples generated using GPT-4 and test data sourced from CoDI-Eval.

\begin{figure}[t]
    \centering
    \includegraphics[scale=0.29]{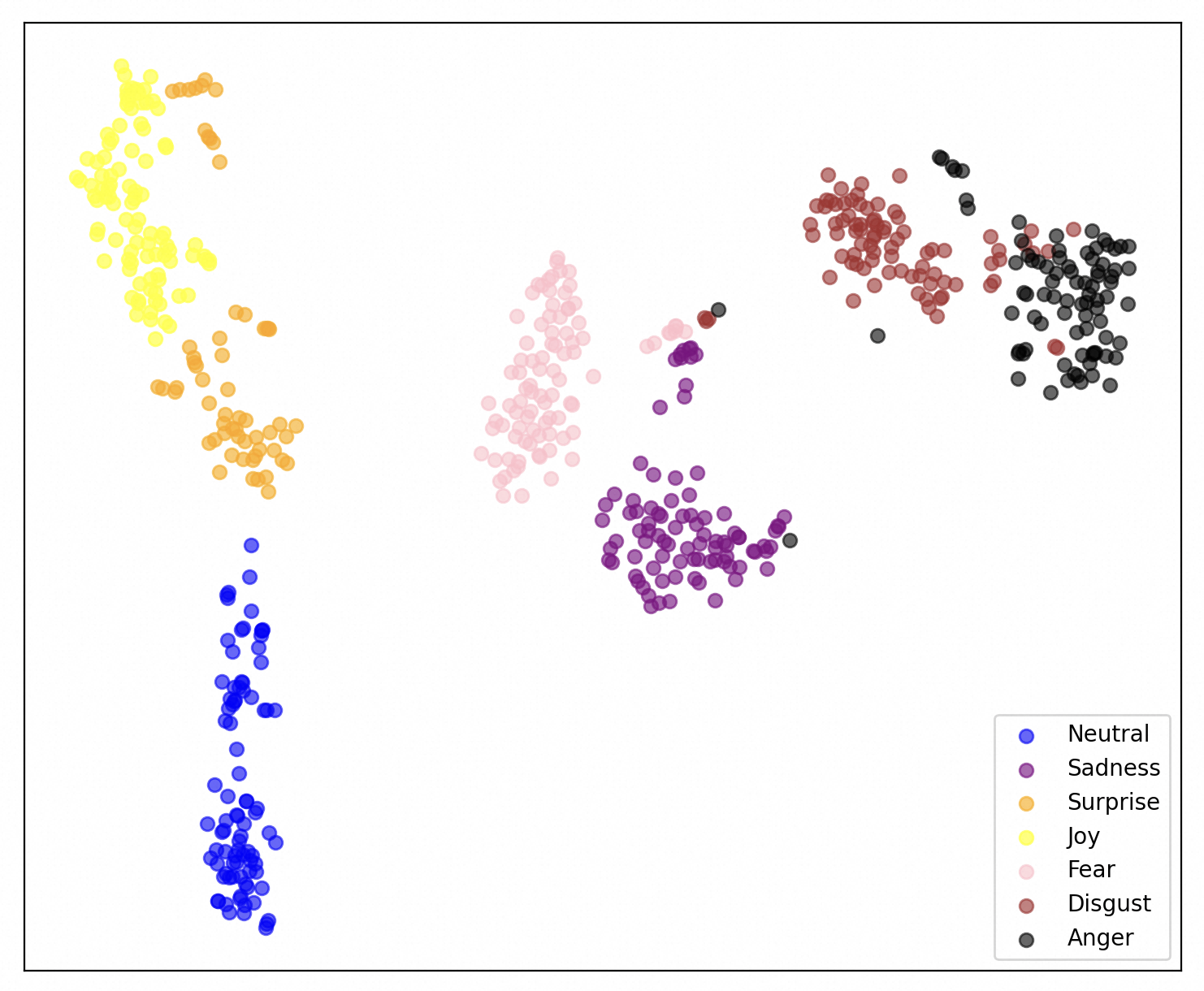}
    \caption{Visualizing the distribution of sentiment attributes.}
    \label{fig:sentiment_distribution}
\end{figure}

\subsection{Evaluation Metrics}\label{appendix:Evaluation metrics}
We use the sentiment\footnote{\url{https://huggingface.co/cardiffnlp/twitter-roberta-base-sentiment-latest}}\textsuperscript{,}\footnote{\url{https://huggingface.co/j-hartmann/emotion-english-roberta-large}} and topic\footnote{\url{https://huggingface.co/cardiffnlp/tweet-topic-21-multi}} RoBERTa classifiers on Huggingface to evaluate sentiment, topic and multi aspects. The detoxification aspect is measured by the Google Perspective API\footnote{\url{https://www.perspectiveapi.com}}. For keyword and length evaluation, we use the following rules: Regarding keywords, we check whether all the required keywords are present in the generated text. If all the keywords are included, it is considered correct. As for length, we evaluate whether the length of the generated text meets the specified label requirements. If it does, it is considered correct.

\section{More experimental results}
\subsection{Visualizing the LoRA weights}
In this section, we visualize the weights of LoRA based on other models, namely Qwen2-72B-Instruct, Llama-3.1-8B/70B-Instruct, and Gemma-2-9B-Instruct, and the results are shown in Figures \ref{fig:qwen2-72b-weight} to \ref{fig:gemma2-weight}. Consistent with the findings in Section \ref{sec:analysis}, the LoRA contribution of each model in the six aspects of Codi-Eval is significantly different, and it can achieve multi-faceted control well.

\subsection{Inputs to the gating function} In our implementation, we use the aspect identifier to input the gating function to learn unique parameters for each aspect and dynamically combine LoRA modules. Comparing this to using hidden states as input, as shown in Table \ref{tab:input to the gating function}, we found that the aspect identifier approach is superior. Using hidden states increases parameter complexity and training difficulty due to the need for a separate gating function for each layer, whereas our approach requires only a global gating function.

\begin{figure}[t]
    \centering
    \includegraphics[scale=0.235]{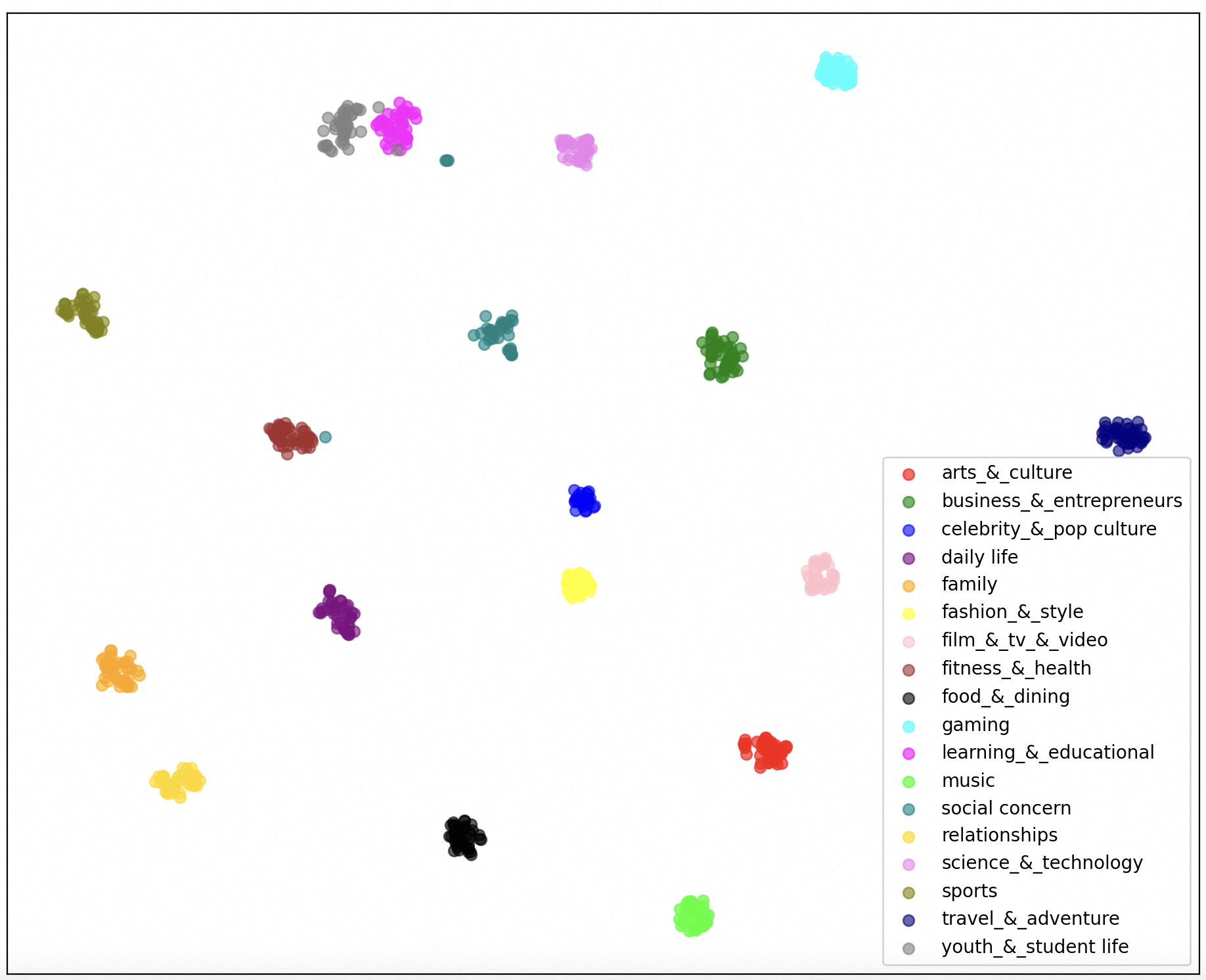}
    \caption{Visualizing the distribution of topic attributes.}
    \label{fig:topic_distribution}
\end{figure}

\subsection{Visualizing attribute distributions}
\label{app:Visualizing attribute distributions}
To validate that our framework effectively achieves attribute awareness by adjusting attribute distributions to reduce conflicts between attributes, we use t-SNE to visualize the distribution of attributes across two aspects: sentiment and topic. Specifically, we feed responses from the attribute set into the classifier. We average the attribute context embeddings for each token in the classifier to obtain sentence-level representations, which are then paired with sentence-level attribute annotations for analysis. The results are presented in Figure \ref{fig:sentiment_distribution} and Figure \ref{fig:topic_distribution}. The figures show that the context embeddings of sentences with different sentiments and topics are separated in the space. This supports the strong attribute-awareness capability of our framework and demonstrates its superior performance in multi-attribute controllability.

\section{Case Study}
\label{app:case study}
We show the cases of LoRA, FFT, and our method on CoDI-Eval in Table \ref{tab:case_see}, and the cases of attribute combinations not seen in CoDI-Eval in Table \ref{tab:case_unsee}.

\begin{table*}[ht]
\centering
\footnotesize 
\renewcommand{\arraystretch}{1.3} 
\begin{tabularx}{\textwidth}{>{\centering\arraybackslash}m{2cm}|m{1cm}|X}
\hline
\multicolumn{1}{c|}{\textbf{Constraints}} & \multicolumn{1}{c|}{\textbf{Method}} & \multicolumn{1}{c}{\textbf{Generation Result}} \\
\hline
\multirow{3}{*}[-3.5ex]{\parbox{2cm}{\centering\textcolor{red}{Surprise}}}  
    & \raisebox{-2.5ex}{LoRA} & In the heart of an ancient forest, where sunlight barely touched the ground, there lay a hidden glade known only to the creatures of the woods. One crisp autumn morning, a curious fox named Faelan stumbled upon this untouched sanctuary. \\
    \cline{2-3}
    & \raisebox{-0.2ex}{FFT} & \textcolor{red}{Wow}, did you really just win the lottery? That's \textcolor{red}{unbelievable}!\\
    \cline{2-3}
    & \raisebox{-1.5ex}{Ours} & \textcolor{red}{I can hardly believe my eyes}! The transformation is incredible, almost \textcolor{red}{magical}. The space looks completely different and \textcolor{red}{better than I ever imagined}. Great job! \\
\hline
\multirow{3}{*}[-8ex]{\parbox{2cm}{\centering\textcolor{red}{Fear}}}  
    & \raisebox{-2.5ex}{LoRA} & As the clock struck midnight, the once bustling city fell eerily silent. Shadows danced in the flickering streetlights, and a cold breeze whispered through empty alleyways. \textcolor{red}{I shivered as footsteps echoed in the distance, slowly approaching.}  \\
    \cline{2-3}
    & \raisebox{-2.5ex}{FFT} & I'm really \textcolor{red}{worried} about the upcoming test. I haven't been able to fully prepare and \textcolor{red}{I'm scared of failing}. Everyone else seems to be more confident, which makes me feel even worse. I hope I can do better than last time, but I'm just so nervous. \\
    \cline{2-3}
    & \raisebox{-3.5ex}{Ours} & I can feel my heart racing as I walk through the darkened streets. Every shadow seems to hide a threat, and every sound feels like it could be a warning sign. \textcolor{red}{The fear of something unknown looms over me}, causing every step to be cautious and deliberate. I know I should feel safe in the light of the moon, but \textcolor{red}{the eerie silence of the night amplifies my deepest fears.} \\
\hline
\multirow{3}{*}[-12ex]{\parbox{2cm}{\centering\textcolor{red}{Disgust}}}  
    & \raisebox{-2.5ex}{LoRA} & I can't believe I have to go through this again. Every time I think about having to deal with such incompetence, my stomach turns. It's absolutely infuriating how these people manage to mess everything up so badly, and yet they think they're doing a good job.  \\
    \cline{2-3}
    & \raisebox{-6.5ex}{FFT} & The 15-year-old boy was charged with attempted murder after he stabbed his classmate in the buttocks with a pen because he believed it would make him popular. The victim is expected to recover, but the boy faces up to 15 years in juvenile detention. The boy’s attorney argued that his client was merely trying to emulate his popular classmates by acting out during a prank gone wrong. However, the judge ruled that the severity of the attack was not justified by any intent to gain popularity, and ordered the boy to face trial as an adult.\\
    \cline{2-3}
    & \raisebox{-3.5ex}{Ours} & I am utterly appalled and \textcolor{red}{repulsed} by the sheer negligence and messiness that surrounds us. It's an absolute disgrace to witness such a chaotic and unkempt situation. \textcolor{red}{The disregard for order and cleanliness is truly revolting.} \\
\hline
\multirow{3}{*}[-6ex]{\parbox{2cm}{\centering\textcolor{blue}{Fashion\&Style}}}  
    & \raisebox{-3.5ex}{LoRA} & \textcolor{blue}{Fashion is not just about wearing trendy clothes, it's about expressing oneself and showcasing individuality through the art of styling}. Whether it's through bold colors, innovative textures, or unique patterns, each outfit tells a story and reflects personal tastes and preferences. \textcolor{blue}{Style goes beyond mere aesthetics; it's an expression of identity and creativity}. \\
    \cline{2-3}
    & \raisebox{-1.5ex}{FFT} & \textcolor{blue}{Embrace the art of fashion}, where each stitch tells a story and every fabric \textcolor{blue}{sings a melody of style}.\\
    \cline{2-3}
    & \raisebox{-1.5ex}{Ours} & \textcolor{blue}{Staying ahead in the ever-evolving world of fashion} requires not only a keen sense of trends but also a personal touch that \textcolor{blue}{makes your style unique and expressive}. \\
\hline
\multirow{3}{*}[-8ex]{\parbox{2cm}{\centering\textcolor{blue}{Family}}}  
    & \raisebox{-2.5ex}{LoRA} & Hey! Just wanted to check in and see how everyone is doing. Missing all of you and can't wait for our next catch-up session. \textcolor{blue}{Let's plan a fun family outing soon}! Take good care and miss those smiles. Love you all. \\
    \cline{2-3}
    & \raisebox{-2.5ex}{FFT} & Hey everyone! Just wanted to check in and see how everyone is doing. It's been too long since we last caught up, and I'd love to hear all the updates. Maybe we can plan a virtual call or a get-together soon? Let me know what you think. Missing all of you and your amazing stories!\\
    \cline{2-3}
    & \raisebox{-2.5ex}{Ours} & Hey everyone, just wanted to share how much I appreciate all of you. \textcolor{blue}{Family is the best and I feel so lucky to have such amazing supporters in my life}. Let's make sure we stay in close touch and keep the love and laughter flowing. Big hugs to all! \\
\hline
\multirow{3}{*}[-15ex]{\parbox{2cm}{\centering\textcolor{blue}{Daily Life}}}  
    & \raisebox{-4ex}{LoRA} & One notable piece of literature that explores the theme of everyday life is “To Kill a Mockingbird” by Harper Lee. This novel provides a poignant portrayal of daily life in a Southern town during the 1930s, delving into issues of racial injustice and moral growth through the eyes of the young protagonist, Scout Finch.\\
    \cline{2-3}
    & \raisebox{-6.8ex}{FFT} & If you’re looking for something more general that encapsulates the essence of daily routines and interactions, you might consider “Babette’s Feast” by Isak Dinesen. This story delves into the lives of a small group of Danish people over many years, highlighting how their simple lives intersect and transform through shared experiences and personal growth. Both of these pieces offer insightful views into the intricacies of everyday life, showcasing how ordinary moments can weave together to form the tapestry of our existence.\\
    \cline{2-3}
    & \raisebox{-6ex}{Ours} & \textcolor{blue}{Daily routines offer a structured framework that helps organize and manage our time efficiently.} These routines often include consistent wake-up times, meals, commuting, work, exercise, and leisure activities. Establishing and maintaining healthy routines can significantly enhance productivity, reduce stress, and foster a sense of stability. \textcolor{blue}{Whether following a strict schedule or adopting a more flexible approach, daily rituals bring order to our lives, boost focus, and empower us to navigate the demands of everyday life with greater ease and confidence.} \\
\hline
\end{tabularx}
\label{tab:sample_table}
\end{table*}

\begin{table*}[ht]
\centering
\footnotesize 
\renewcommand{\arraystretch}{1.3} 
\begin{tabularx}{\textwidth}{>{\centering\arraybackslash}m{2cm}|m{1cm}|X}
\hline
\multicolumn{1}{c|}{\textbf{Constraints}} & \multicolumn{1}{c|}{\textbf{Method}} & \multicolumn{1}{c}{\textbf{Generation Result}} \\
\hline
\multirow{3}{*}[-10ex]{\parbox{2cm}{\centering\textcolor{red}{Surprise}\\ \textcolor{blue}{Health}}}  
    & \raisebox{-2.5ex}{LoRA} & Did you know? Simply by taking just 30 minutes to exercise each day can dramatically \textcolor{blue}{boost your energy levels, improve your mood, and significantly reduce your risk of chronic diseases}! Don’t wait any longer to transform your health – the time to start is now! \\
    \cline{2-3}
    & \raisebox{-2.5ex}{FFT} & \textcolor{red}{Get ready to shock your system into shape!} Discover the latest trend that's \textcolor{blue}{taking the health world by storm}, promising transformative results in just weeks. Stay tuned for an exclusive sneak peek! \\
    \cline{2-3}
    & \raisebox{-5ex}{Ours} & Wow, who knew that simply swapping your morning coffee for green tea could make such a huge difference in your fitness journey? Just last week, I read about a study that revealed \textcolor{blue}{drinking green tea increases metabolism and boosts fat burning!} Curious, I decided to give it a try. Four days in, I stepped on the scale and—wow, a full 5 pounds lighter! \textcolor{red}{I'm stunned and excited} to see where this change in habit will lead. \textcolor{red}{Green tea, you've surprised me big time!} \\
\hline
\multirow{3}{*}[-10ex]{\parbox{2cm}{\centering\textcolor{red}{Anger}\\ \textcolor{blue}{Food}}}  
    & \raisebox{-2.5ex}{LoRA} & I can't believe how utterly disappointing this meal turned out to be! When I ordered it, I had high hopes, \textcolor{blue}{expecting a delightful feast that would satisfy my palate and energize my day}. But what was served was nothing short of a culinary disaster. \\
    \cline{2-3}
    & \raisebox{-4ex}{FFT} & \textcolor{blue}{Dining out is supposed to be an enjoyable experience}, but it can quickly turn sour when the service is subpar. \textcolor{red}{It's infuriating when you're greeted with indifference and your requests are met with apathy.} The whole point of going out to eat is to relax and enjoy a good meal, not to deal with rude or unattentive staff. \\
    \cline{2-3}
    & \raisebox{-4ex}{Ours} & I can't even begin to express how \textcolor{red}{infuriating} it is to have \textcolor{blue}{dinner ruined by inconsiderate individuals who have absolutely no regard for polite manners or the effort put in by hardworking chefs}. It's \textcolor{red}{exasperating} to see people chattering away, elbowing each other across the table, or blatantly ignoring the obviously uncomfortable glances of their neighbors. \\
\hline
\multirow{3}{*}[-21ex]{\parbox{2cm}{\centering\textcolor{red}{Neutral}\\ \textcolor{blue}{Culture}}}  
    & \raisebox{-6.5ex}{LoRA} & \textcolor{blue}{The Japanese tea ceremony, or Chanoyu, is a centuries-old ritual rooted in harmony, respect, purity, and tranquility. Introduced from China by Buddhist monks in the 9th century, it evolved to reflect Japanese culture and philosophy.} This artful practice involves precise tea preparation, room decoration, and host-guest interactions, fostering mindfulness and connection. The Japanese tea ceremony serves as a reflection of the country's rich cultural heritage and continues to be an integral part of Japanese social and cultural life today. \\
    \cline{2-3}
    & \raisebox{-7ex}{FFT} & \textcolor{red}{In Japan, it is common to find shops, especially those traditional ones, closing their doors by late in the evening.} This is because the Japanese typically prefer to shop during the day, and many establishments adhere to regular business hours that reflect this. \textcolor{red}{In fact, many cities in Japan have laws regulating store hours to ensure that streets are not too crowded, maintaining a peaceful atmosphere in the evenings.} This tradition underscores the value the Japanese place on maintaining a balanced lifestyle that separates work from personal time, promoting a serene evening environment. \\
    \cline{2-3}
    & \raisebox{-8ex}{Ours} & \textcolor{blue}{The Japanese tradition of "chado" or Japanese tea ceremony, is a refined and ritualistic way of preparing and serving tea.} Dating back to the 9th century, it is considered a spiritual art form that emphasizes mindfulness, respect, and purity. \textcolor{red}{During a tea ceremony, participants engage in a series of meticulous steps involving the preparation of matcha tea, a powdered green tea. The atmosphere is quietly serene, and the surroundings are often adorned with delicate art pieces which enhance the aesthetic appeal.} This cultural practice not only serves the purpose of enjoying tea but also fosters a meditative and reflective experience for both the participants and the host. \\
\hline
\multirow{3}{*}[-1ex]{\parbox{2cm}{\centering\textcolor{olive}{At Most \\ 10 Words}}}  
    & LoRA & She savored the warm, spiced aroma of cinnamon rolling through the air. (12 words) \\
    \cline{2-3}
    & \raisebox{-1.5ex}{FFT} & Exploring new cities enriches life, offering unique experiences and broadening perspectives. (11 words) \\
    \cline{2-3}
    & Ours & \textcolor{olive}{The cat slept peacefully on the windowsill.} (7 words) \\
\hline
\multirow{3}{*}[-7ex]{\parbox{2cm}{\centering\textcolor{olive}{Around \\ 32 to 40 Words}}}  
    & \raisebox{-2.5ex}{LoRA} &\textcolor{olive}{Autumn brings a vibrant change to the landscape, with leaves transforming into a mosaic of reds, oranges, and yellows, creating a serene and picturesque scene that delights both the eyes and soul.} (32 words) \\
    \cline{2-3}
    & \raisebox{-2.5ex}{FFT} & Beneath the silvery moon, whispers danced through the ancient trees, each leaf murmuring secrets of old, weaving tales of forgotten times where magic and reality blurred at the forest's edge. (30 words) \\
    \cline{2-3}
    & \raisebox{-2.5ex}{Ours} & \textcolor{olive}{Exploring new hobbies can significantly enhance one's creativity and mental well-being. Whether it's painting, cooking, or learning a musical instrument, each activity offers a unique outlet for stress and a way to express oneself.} (34 words) \\
\hline

\end{tabularx}
\label{tab:sample_table}
\end{table*}

\begin{table*}[ht]
\centering
\footnotesize 
\renewcommand{\arraystretch}{1.3} 
\begin{tabularx}{\textwidth}{>{\centering\arraybackslash}m{2cm}|m{1cm}|X}
\hline
\multicolumn{1}{c|}{\textbf{Constraints}} & \multicolumn{1}{c|}{\textbf{Method}} & \multicolumn{1}{c}{\textbf{Generation Result}} \\
\hline
\multirow{3}{*}[-14ex]{\parbox{2cm}{\centering\textcolor{olive}{Equal to \\ 
60 Words}}}  
    & \raisebox{-5ex}{LoRA} & Exploring the art of precise language demands meticulous attention to detail, ensuring each word serves a purpose. Crafting sentences with exactness mirrors the meticulousness of an artist, where every element is deliberate and impactful. This skill not only enhances clarity but also enriches the reader's experience, offering a concise yet profound narrative that resonates on multiple levels. (57 words) \\
    \cline{2-3}
    & \raisebox{-5ex}{FFT} & The art of crafting concise prose is both challenging and rewarding. Every word must be chosen with care, ensuring it contributes meaningfully to the overall message. This exercise in brevity helps sharpen one's thinking and enhances the ability to communicate effectively. Precisely measured in length, the text not only fits a specific format but also tests the writer's skill in being succinct yet expressive. (64 words) \\
    \cline{2-3}
    & \raisebox{-5ex}{Ours} & \textcolor{olive}{The serene landscape beckoned, with lush green fields stretching towards the horizon. A lone tree stood majestically, offering shade and solace. Birds chirped melodiously, creating a symphony of natural sounds. This picturesque scene was a perfect escape from the hustle and bustle of daily life, invoking peace and tranquility within the observer's soul. Such moments are precious and truly cherishable.} (60 words) \\
\hline
\multirow{3}{*}[-2.5ex]{\parbox{2cm}{\centering\textcolor{purple}{"clothe"\\ "mother" \\"toddler"}}}  
    & \raisebox{-1.5ex}{LoRA} &The \textcolor{purple}{mother} quickly \textcolor{purple}{clothed} the \textcolor{purple}{toddler} in a warm jacket before heading out to play in the snow. \\
    \cline{2-3}
    & FFT & The \textcolor{purple}{mother} dressed her \textcolor{purple}{toddler} in a cute little \textcolor{purple}{clothe} that matched her vibrant personality. \\
    \cline{2-3}
    & \raisebox{-1.5ex}{Ours} & While shopping at the store, my \textcolor{purple}{mother} noticed a colorful \textcolor{purple}{toddler} in the aisle trying on various \textcolor{purple}{clothes}, sparking a fun afternoon of play pretend. \\
\hline
\multirow{3}{*}[-15ex]{\parbox{2cm}{\centering\textcolor{purple}{"demonstrate"\\ "neck" \\"tie"}}}  
    & \raisebox{-6ex}{LoRA} &At the annual company gala, every team member looked sharp in formal attire, meticulously grooming their appearances to impress. John, always keen on details, decided to \textcolor{purple}{demonstrate} his newfound interest in fashion by choosing a silk \textcolor{purple}{tie} with a subtle pattern, which complemented his suit perfectly. He also made sure to adjust the collar and of his shirt multiple times to ensure he was camera-ready for the speeches and photographs that would follow. (\textcolor{purple}{lack neck}) \\
    \cline{2-3}
    & \raisebox{-6ex}{FFT} & During his presentation, James decided to \textcolor{purple}{demonstrate} the proper technique for tying a knot. He carefully adjusted his necktie, ensuring each fold was precise and symmetrical as he showed the audience how to achieve a perfect knot. The room watched attentively, eager to learn this skill that would elevate their appearance. As James finished his demonstration, he adjusted his own \textcolor{purple}{tie} one last time before taking his seat, satisfied with the clarity and gracefulness of his display. (\textcolor{purple}{lack neck}) \\
    \cline{2-3}
    & \raisebox{-5ex}{Ours} & During his first day at the new job, Jacob wanted to make a strong impression with his professionalism. He decided to \textcolor{purple}{demonstrate} his commitment by choosing a sharp, classic blue suit. He carefully adjusted the \textcolor{purple}{tie} around his \textcolor{purple}{neck}, ensuring it was symmetrical and presented a polished look. Jacob knew that attention to detail in his attire could set the tone for his entire career. \\
\hline
\multirow{3}{*}[-17ex]{\parbox{2cm}{\centering\textcolor{purple}{"dance"\\ "music"\\ "perform"\\ "room"}}}  
    & \raisebox{-6ex}{LoRA} &The grand \textcolor{purple}{room} was transformed for the evening's performance, draped in twinkling lights and adorned with eager spectators. As the \textcolor{purple}{music} began to fill the air, a mesmerizing \textcolor{purple}{dance} commenced, showcasing the artist's mastery over their craft. Each note played in perfect harmony with the movements, captivating everyone in the space and creating an electric atmosphere of artistry and entertainment. Performers and audience alike were swept up in the joy and creativity unfolding before them. (\textcolor{purple}{lack perform})\\
    \cline{2-3}
    & \raisebox{-7.5ex}{FFT} & In the heart of the bustling city, the local community center decided to host a vibrant evening of cultural enrichment. They arranged for a spectacular \textcolor{purple}{dance} performance in the main room, where aspiring dancers and seasoned performers alike would take the stage to showcase their unique talents. The \textcolor{purple}{room} was filled with anticipation, as the sounds of compelling \textcolor{purple}{music} echoed through the space, setting the tone for an unforgettable night. This event not only celebrated the diverse artistic abilities within the community but also fostered a spirit of unity and joy among the attendees. (\textcolor{purple}{lack perform}) \\
    \cline{2-3}
    & \raisebox{-5ex}{Ours} & The \textcolor{purple}{room} buzzed with excitement as the students prepared to \textcolor{purple}{perform} their carefully choreographed \textcolor{purple}{dance}. The \textcolor{purple}{music} filled the air, elevating the spirits of everyone present. As the first notes played, their hard work and dedication were brought to life through their graceful movements and impeccable timing. It was an evening where artistry flourished, and the walls of the classroom turned into a stage, hosting a spectacle of talent and passion.\\
\hline
\end{tabularx}
\label{tab:sample_table}
\end{table*}

\begin{table*}[ht]
\centering
\footnotesize 
\renewcommand{\arraystretch}{1.3} 
\begin{tabularx}{\textwidth}{>{\centering\arraybackslash}m{2cm}|m{1cm}|X}
\hline
\multicolumn{1}{c|}{\textbf{Constraints}} & \multicolumn{1}{c|}{\textbf{Method}} & \multicolumn{1}{c}{\textbf{Generation Result}} \\
\hline
\multirow{3}{*}[-8ex]{\parbox{2cm}{\centering\textbf{Nontoxicity}}}  
    & \raisebox{-2.5ex}{LoRA} &I appreciate the feedback, Eley. Let's definitely find time to discuss this further and explore different perspectives on the topic. It's always great to engage in thoughtful conversations that can enhance our understanding. \\
    \cline{2-3}
    & \raisebox{-2.5ex}{FFT} & I'm really sorry about my previous remarks, Eley. I didn't mean to offend anyone and I'm open to discussing this further if you'd like to clear things up. Let's find a time that works for you to talk things through. \\
    \cline{2-3}
    & \raisebox{-4ex}{Ours} & Meminist shirts are just making fun of feminism because the people who wear them don't give a fair chance to understand the genuine concerns and goals of the feminist movement. It's important to engage in open and respectful dialogues that can help bridge the gap between different perspectives and promote a more inclusive society for all. \\
\hline
\multirow{3}{*}[-8ex]{\parbox{2cm}{\centering\textbf{Nontoxicity}}}  
    & \raisebox{-1.5ex}{LoRA} &The male fled the scene on a bicycle, but not before he shouted "I'll get you next time!" His tone was filled with a mix of frustration and defiance as he pedaled away swiftly. \\
    \cline{2-3}
    & \raisebox{-2.5ex}{FFT} & On Friday, Trump called these people sons of hardworking Americans who deserve to have their voices heard and their concerns addressed. He emphasized the importance of listening to all citizens and working together to find solutions that benefit everyone in society. \\
    \cline{2-3}
    & \raisebox{-5.5ex}{Ours} & She saw the curve of her hips and the shape of her waist reflected back at her in the mirror. With a smile, she admired how her outfit complemented her figure, enhancing her natural beauty. Today, she felt confident and ready to take on the challenges ahead. As she left the house, she carried with her not just her belongings, but also a positive mindset, eager to make the most out of every opportunity that came her way. \\
\hline
\multirow{3}{*}[-8ex]{\parbox{2cm}{\centering\textbf{Nontoxicity}}}  
    & \raisebox{-5.5ex}{LoRA} &I'd happily do the jail time if I could get just one solid punch in to the face of someone who truly deserves it, but violence isn't the answer. Instead, I choose to channel my anger and frustration into positive actions that can create real change and make a difference in the world. By advocating for justice and understanding, we can work towards a more equitable society where everyone has the opportunity to thrive. \\
    \cline{2-3}
    & \raisebox{-2.5ex}{FFT} &The Miami Herald reports Mohamed T. had a note in his pocket with phrases in both Arabic and German including "I want to kiss a flower that blooms only at midnight under the full moon". \\
    \cline{2-3}
    & \raisebox{-2.5ex}{Ours} &He looked me directly in the eye and drunkenly said, "I'm gonna shake your hand, because tonight has been amazing!" He then proceeded to extend his hand with a wide, enthusiastic smile, ready to seal our new friendship with a firm handshake. \\
\hline
\end{tabularx}
\caption{Example cases of generated sentences in the CoDI-Eval test set. The above cases are generated based on Qwen2-7B-Instruct. \textcolor{red}{Red} text highlights content related to sentiment. \textcolor{blue}{Blue} text highlights content related to topic. \textcolor{olive}{Olive} text highlights content related to length. \textcolor{purple}{Purple} text highlights content related to keywords.}
\label{tab:case_see}
\end{table*}

\begin{table*}[ht]
\centering
\footnotesize 
\renewcommand{\arraystretch}{1.3} 
\begin{tabularx}{\textwidth}{>{\centering\arraybackslash}m{2cm}|m{1cm}|X}
\hline
\multicolumn{1}{c|}{\textbf{Constraints}} & \multicolumn{1}{c|}{\textbf{Method}} & \multicolumn{1}{c}{\textbf{Generation Result}} \\
\hline
\multirow{3}{*}[-14.5ex]{\parbox{2cm}{\centering\textcolor{blue}{sports} \\ \textcolor{purple}{"playground"\\ "breakfast" \\"sky"}}}  
    & \raisebox{-9ex}{LoRA} &In the heart of the bustling city stood an old, weathered \textcolor{purple}{playground}, a place where children's laughter echoed under the clear blue \textcolor{purple}{sky}. One particular sunny day, the air crisp and inviting, the children decided to have a friendly soccer match. They brought along a homemade \textcolor{purple}{breakfast} of sandwiches and fresh fruits, spreading a checkered cloth on the picnic table under the shade of an old oak tree. As they ate, their chatter filled the air, weaving stories and dreams. \textcolor{blue}{The game started, and the children played with an enthusiasm that only a playground can inspire. With every goal, their joy under the vast sky grew, creating memories that would last a lifetime.} \\
    \cline{2-3}
    & \raisebox{-2.5ex}{FFT} & Under the bright morning \textcolor{purple}{sky}, families gathered at the \textcolor{purple}{playground}, enjoying a quick \textcolor{purple}{breakfast} before the games began. \textcolor{blue}{Kids raced from the swings to the field, laughter echoing as parents cheered them on, coffee cups in hand, ready for a day of friendly competition.} \\
    \cline{2-3}
    & \raisebox{-4ex}{Ours} & As the early morning \textcolor{purple}{sky} turned a brilliant shade of blue, the \textcolor{purple}{playground} buzzed with excitement. Families gathered for \textcolor{purple}{breakfast} under the rising sun, kids munching on snacks before they rushed off to play. \textcolor{blue}{A friendly soccer match soon broke out, filling the air with cheers as everyone enjoyed a perfect sports-filled morning.} \\
\hline
\multirow{3}{*}[-11.5ex]{\parbox{2cm}{\centering\textcolor{olive}{Around \\ 50 to 70 words} \\ \textcolor{purple}{"singer"\\ "money" \\"company"}}}  
    & \raisebox{-3.5ex}{LoRA} &\textcolor{olive}{A renowned} \textcolor{purple}{singer} \textcolor{olive}{recently invested in a new tech} \textcolor{purple}{company} \textcolor{olive}{that specializes in music streaming services. The goal is to generate more revenue and connect directly with fans without intermediaries taking a large cut of his} \textcolor{purple}{money}\textcolor{olive}{. This venture might open new pathways for artists to earn substantially, influencing the industry's revenue models.} (53 words)\\
    \cline{2-3}
    & \raisebox{-5ex}{FFT} & \textcolor{olive}{The rising star, now a renowned} \textcolor{purple}{singer}\textcolor{olive}{, decided to invest in a new tech} \textcolor{purple}{company}\textcolor{olive}{. With her increasing fortune, she aimed to revolutionize the music industry by integrating cutting-edge technology to enhance artists' performances. This strategic venture promised not only to boost her career but also to transform the way music is experienced globally.} (\textcolor{purple}{lack money}, 54 words)\\
    \cline{2-3}
    & \raisebox{-5ex}{Ours} & \textcolor{olive}{The young} \textcolor{purple}{singer} \textcolor{olive}{signed a deal with a music} \textcolor{purple}{company}\textcolor{olive}{, hoping it would be her big break. She knew that fame could bring not only recognition but also} \textcolor{purple}{money} \textcolor{olive}{to support her dreams. Despite the challenges in the industry, she stayed focused, writing songs late into the night. The company promised promotions, and with every note she sang, she felt her future growing brighter.} (64 words) \\
\hline
\multirow{3}{*}[-15ex]{\parbox{2cm}{\centering\textcolor{red}{positive} \\ \textcolor{olive}{Around \\ 50 to 70 words} \\ \textcolor{purple}{"beautiful"\\ "success"}}}  
    & \raisebox{-5.2ex}{LoRA} &\textcolor{red}{Embrace each day with optimism, for every moment is a chance to create} \textcolor{purple}{beautiful} \textcolor{red}{memories}. \textcolor{olive}{Whether small victories or significant} \textcolor{purple}{success}\textcolor{olive}{, each step forward enriches our journey. Cherish these moments, for they beautifully weave the tapestry of our lives, full of joy and accomplishment.} \textcolor{red}{Keep striving and remember, every success is a spark of beauty in the endless horizon of our endeavors.} (63 words)\\
    \cline{2-3}
    & \raisebox{-5.1ex}{FFT} & \textcolor{olive}{Achieving} \textcolor{purple}{success} \textcolor{olive}{is truly} \textcolor{purple}{beautiful}\textcolor{olive}{. It's a feeling that fills you with pride and joy, giving you a sense of fulfillment. Whether it's personal or professional milestones,} \textcolor{red}{each victory is a stepping stone to greater heights. Let's cherish these moments and continue to strive for happiness and success in our endeavors. Remember, every small step forward is a beautiful part of your journey.} (63 words)\\
    \cline{2-3}
    & \raisebox{-5ex}{Ours} & \textcolor{olive}{The day was} \textcolor{purple}{beautiful}\textcolor{olive}{, with the sun shining brightly and a gentle breeze lifting everyone's spirits. Jane felt a sense of} \textcolor{purple}{success} \textcolor{olive}{as she completed her project, something she'd worked on tirelessly for weeks.} \textcolor{red}{Her friends gathered to celebrate, reminding her that hard work truly pays off. It was a moment to savor, filled with joy, laughter, and a beautiful sense of accomplishment.} (63 words)\\
\hline
\end{tabularx}
\caption{Examples of multi-aspect generation involving combinations of aspects not seen in CoDI-Eval. The above cases are generated based on Llama-3.1-8B-Instruct.}
\label{tab:case_unsee}
\end{table*}

\end{document}